\documentclass[11pt]{article}

\usepackage[T1]{fontenc}
\usepackage{lmodern}
\usepackage[margin=1in]{geometry}
\usepackage{amsmath,amssymb,amsfonts,mathtools}
\usepackage{bm}
\usepackage[ruled,vlined]{algorithm2e}
\usepackage{graphicx}
\usepackage{booktabs}
\usepackage{array}
\usepackage{tabularx}
\usepackage{enumitem}
\usepackage{xcolor}
\usepackage{microtype}
\usepackage{float}
\usepackage[section]{placeins}        
\usepackage{caption}
\usepackage[square,numbers,sort&compress]{natbib}
\usepackage{hyperref}
\usepackage{authblk}


\hypersetup{
    colorlinks=true,
    linkcolor=blue!60!black,
    citecolor=blue!60!black,
    urlcolor=blue!60!black,
    pdftitle={FlowersBench: A FluidFlower-Inspired Dataset for Benchmarking Carbon Storage Surrogates Across Geological Structures},
    pdfauthor={Farah Rabie, Hannah Menke, Florian Doster, Amy Gough, Ahmed H. Elsheikh}
}

\title{\textit{FlowersBench}: A FluidFlower-Inspired Dataset for Benchmarking\\ Carbon Storage Surrogates Across Geological Structures}

\author[1,*]{Farah Rabie}
\author[1]{Hannah Menke}
\author[1]{Florian Doster}
\author[1]{Amy Gough}
\author[1,*]{Ahmed H. Elsheikh}
\affil[1]{Institute of GeoEnergy Engineering (IGE), \protect\\School of Energy, Geoscience, Infrastructure \& Society, \protect\\Heriot-Watt University, Edinburgh, EH14 4AS, UK}
\date{}

\begin{document}
\maketitle
\begin{NoHyper}
{\renewcommand{\thefootnote}{*}\footnotetext{Corresponding authors: \href{mailto:f.rabie@hw.ac.uk}{f.rabie@hw.ac.uk}, \href{mailto:a.elsheikh@hw.ac.uk}{a.elsheikh@hw.ac.uk}}}
\end{NoHyper}

\begin{abstract}

Machine learning surrogates are emerging as a computationally efficient tool for uncertainty quantification in geological carbon storage (GCS). Geological structure constitutes a major source of uncertainty; however, existing benchmark simulation datasets predominantly vary petrophysical and operational parameters within a fixed geological setting. Generalisation to unseen geological structures therefore remains largely unexplored. We introduce \textit{FlowersBench}, a benchmark dataset of 5,000 two-dimensional, reservoir-scale CO$_2$ storage simulations across 244 structural geometries. The geometries are sampled from an in-house structural generator inspired by the FluidFlower experiments and Case B of the 11th Society of Petroleum Engineers Comparative Solution Project (SPE11B). Each simulation is run with Open Porous Media (OPM) Flow over a 50-year injection period and a 500-year post-injection monitoring period. Test geometries are withheld from training to assess generalisation to unseen structural geometries. Two additional subsets support evaluation under structural shift. The first extends the parameter ranges of the structural generator beyond those used for training, and the second introduces an unseen topology of multiple parallel faults. A baseline U-Net surrogate is trained to predict CO$_2$ gas saturation. In representative test cases, the surrogate recovers the main migration pathways, including lateral spreading beneath low-permeability layers and upward flow along fault damage zones, with errors concentrated near the plume boundary. Under both structural shifts, prediction error exceeds that on the held-out geometries, and the increase in relative error is greater for the unseen fault topology. \textit{FlowersBench} is publicly available and provides a common basis for evaluating GCS surrogates across geological structures.

\end{abstract}

\section{Introduction} \label{sec:introduction}

Geological carbon storage (GCS) is widely considered a key strategy for large-scale decarbonisation \citep{ipcc-2022, macdowell-2018, metz-2005}. Deploying GCS at scale relies on numerical simulations of multiphase flow to evaluate candidate storage sites and injection strategies \citep{celia-2015}. These simulations must resolve processes across a wide range of spatial and temporal scales, from near-well plume migration to far-field pressure propagation, and are therefore computationally demanding \citep{celia-2015, bandilla-2015}. The computational cost is particularly limiting for forward uncertainty quantification, which requires repeated simulations across many scenarios to characterise the effects of geological and operational uncertainty on storage-system behaviour \citep{pawar-2015, lu-2024}.

Machine learning (ML) surrogates offer a computationally efficient approximation to numerical reservoir simulators. Trained on simulation data, these models learn mappings from geological and operational inputs, such as permeability and porosity fields or well locations, to quantities of interest such as gas saturation and pressure fields \citep{wen-2022, wen-2023, ju-2024, badawi-2025}. Once trained, surrogates predict these quantities for new scenarios at a fraction of the simulation cost, making large-scale uncertainty quantification computationally tractable \citep{wen-2022, lu-2024}. In practice, surrogates may be applied to configurations absent from the training data, including storage sites with geological structures unlike those represented during training. Accuracy on held-out samples drawn from the same distribution as the training data does not by itself establish how well a surrogate will perform in these settings \citep{egele-2024}.

Generalisation across geological structures is particularly important. Subsurface geometry is a major source of uncertainty at storage sites and is often impossible to determine uniquely from seismic and borehole observations \citep{bond-2015}. \citet{bond-2007} demonstrated this challenge by analysing 412 interpretations of a synthetic seismic image generated from a known subsurface structure; only 21\% of participants recovered the intended geological concept. Uncertainty in predicted storage behaviour also persists when the geometry is known. In the FluidFlower validation benchmark, nine research groups forecast CO$_2$ migration in the same well-characterised laboratory geometry. Their predictions of plume evolution differed substantially, owing partly to different modelling choices and interpretations of petrophysical properties \citep{fluidflower-2024}. At field scale, uncertainty in the geometry itself further complicates reliable prediction of storage behaviour.

Surrogate generalisation is commonly assessed in two settings. Generalisation to unseen data drawn from the training distribution is referred to as interpolation, whereas generalisation to data from a shifted distribution is referred to as extrapolation or out-of-distribution (OOD) generalisation. In this work, we refer to evaluation on unseen geometries drawn from the training distribution as in-distribution evaluation, and to evaluation on geological structures that differ systematically from those represented during training as evaluation under structural shift.

Existing benchmark simulation datasets for GCS surrogates predominantly evaluate in-distribution generalisation along petrophysical and operational axes while holding geological structure fixed. \citet{wen-2022} represent permeability and porosity as heterogeneous random fields on a two-dimensional, radially symmetric domain, together with scalar injection and reservoir parameters, and evaluate a U-Net-enhanced Fourier neural operator (U-FNO) on held-out samples from the same field distribution. \citet{wen-2023} extend this framework to three dimensions and multiple wells by varying dip angle, depth, and permeability statistics, but retain a single layered, unfaulted reservoir structure throughout training and evaluation. \citet{badawi-2025} vary well location, well number, and operational controls, evaluating predictions on withheld well configurations, while representing petrophysical uncertainty through two-point geostatistics within a fixed reservoir geometry. Across these studies, petrophysical, well-configuration, and operational variability is tested extensively, but stratigraphic and fault architecture is not.

Where faults are included, existing studies vary fault properties while keeping fault geometry fixed. \citet{ju-2024} explicitly incorporate faults and show that a graph neural network generalises across meshes with different discretisations, well locations, and permeability fields, with fault position and domain shape unchanged across all realisations. \citet{lu-2024} vary fault permeability, in-situ stress ratio, and rock compressibility through a space-filling design for a coupled flow-geomechanics surrogate, while retaining a fixed fault geometry. These variations represent uncertainty in fault properties rather than in geological structure, leaving structural uncertainty largely outside the scope of the benchmark simulation datasets reviewed here.

Related work in multiphase-flow surrogate modelling suggests that including geometric diversity in the training data can improve generalisation to unseen geometries. At the pore scale, \citet{abdellatif-2026} found that models trained across multiple levels of grain-packing heterogeneity achieved lower errors on a held-out heterogeneity level than models trained on a single level. Although this study considers a discrete parametric sweep rather than structural variability at reservoir scale, it supports the inclusion of geometric diversity in surrogate benchmarks.

Here, we introduce \textit{FlowersBench}, a benchmark dataset for assessing how GCS surrogates generalise across geological structures, both to unseen geometries and under controlled structural shifts. A parametrised, in-house structural generator varies stratigraphic layering and fault configurations while preserving the layered architecture, fault-controlled compartmentalisation, and bounded domain of the FluidFlower configuration \citep{fluidflower-2024, nordbotten-2024}. The parameter ranges of the generator are chosen to produce plausible and diverse geometries consistent with this analogue. The benchmark therefore evaluates generalisation for a defined synthetic structural family rather than against a field-derived uncertainty distribution.

\textit{FlowersBench} is built on 244 structural geometries sampled from the in-house structural generator. Within each geometry, we sample several petrophysical property realisations and injection-well locations. Each simulation case pairs one property realisation with one well location, giving 5,000 cases in total. Every case is simulated with Open Porous Media (OPM) Flow (release 2025.04) \citep{rasmussen-2020} on a two-dimensional cross-section of 8.40 km $\times$ 1.20 km, over a 50-year injection period followed by a 500-year post-injection monitoring period. Reflecting each case about the vertical axis, following \citet{badawi-2025}, doubles the dataset to 10,000 samples without additional simulation.

We use \textit{FlowersBench} to evaluate a baseline U-Net \citep{ronneberger-2015} surrogate that predicts CO$_2$ gas saturation at 17 reporting times. For in-distribution evaluation, the data are split by structural geometry, and each geometry and the reflected samples derived from it are assigned exclusively to one of the training, validation, or test sets. For evaluation under structural shift, two further test subsets are generated with the same workflow, subject to the modifications described in Section \ref{sec:surrogate_extrapolation_results}. The first extends selected parameter ranges of the structural generator beyond those used for training while retaining the training fault topologies. The second introduces an unseen topology of multiple parallel faults that the training generator cannot produce under any parameter setting. These subsets are defined through controlled changes to the generation process rather than through a statistical measure of distance from the training data. This evaluation is therefore described as structural shift rather than OOD generalisation. Performance is reported using geometry-balanced, plume-masked metrics. The baseline demonstrates how the benchmark is used and provides a reference for future comparison. It is not intended to establish how well GCS surrogates generalise across geological structures, a question that \textit{FlowersBench} is designed to support rather than resolve.

The remainder of this paper is organised as follows. Section \ref{sec:background} introduces the geological and physical background, including the SPE11B benchmark on which the simulation setup is based. Section \ref{sec:methodology} describes how the dataset is constructed, and Section \ref{sec:dataset_characterisation} characterises the diversity of the resulting cases. Section \ref{sec:surrogate_baseline} uses the dataset to train and evaluate the U-Net surrogate, both in-distribution and under structural shift. Section \ref{sec:discussion} discusses what these results imply for surrogate benchmarking, and Section \ref{sec:limitations} outlines the limitations of the benchmark and directions for future work.

\section{Background} \label{sec:background}

Geological structure, defined by the arrangement and hydraulic contrasts of stratigraphic layers and faults, governs the migration and retention of injected CO$_2$. Buoyant CO$_2$ rises until it encounters a sealing layer, beneath which it accumulates when the capillary entry pressure of the seal exceeds the buoyancy pressure of the plume \citep{jayasekara-2020}. This structural trapping therefore depends on the geometry and continuity of sealing layers, with variations in layering and faulting redirecting plume migration and altering the volume of CO$_2$ retained beneath a seal. Residual, solubility, and mineral trapping also contribute to long-term CO$_2$ retention. Residual trapping immobilises disconnected CO$_2$ ganglia through capillary snap-off behind the migrating plume \citep{sun-2020}. Dissolution transfers CO$_2$ into the aqueous phase and can be enhanced by density-driven convective mixing \citep{luo-2022}, while geochemical reactions with the host rock can convert dissolved CO$_2$ into stable carbonate minerals \citep{wu-2021}. Over the timescales considered here, free-phase CO$_2$ migration and distribution are strongly influenced by geological structure, making structural variability a key consideration in evaluating surrogate predictions.

Faults add further structural complexity, displacing sealing layers and altering the connectivity of pathways available for fluid migration. Fault throw and dip determine which layers are juxtaposed across a fault. The properties of the fault zone control whether the fault acts as a barrier or as a conduit. For example, clay smear along a fault can impede flow, whereas discontinuities in the smear allow fluids to cross \citep{vrolijk-2016}. Hydraulic properties may also vary substantially along a single fault, creating localised leakage paths within otherwise sealing structures \citep{salo-salgado-2023}. Both fault geometry and fault hydraulic properties are difficult to constrain uniquely from seismic and borehole data \citep{bond-2015, salo-salgado-2023}. To represent this uncertainty in simplified form, \textit{FlowersBench} samples fault throw and dip and assigns separate hydraulic properties to fault cores and damage zones, with cores classified as sealing or non-sealing.

More broadly, structural uncertainty is commonly represented in geological modelling through ensembles of models that sample alternative interpretations of the subsurface \citep{wellmann-2018}. \citet{tacher-2006} assigned probability distributions to stratigraphic boundaries and nonconformities, propagating the resulting uncertainty from sparse cross-sections and borehole observations. \citet{wellmann-2010} generated alternative structural models through Monte Carlo perturbation of structural measurements, such as layer orientations and contact locations. In reservoir engineering, structural uncertainty has likewise been propagated into probabilistic estimates of gross rock volume, hydrocarbons in place, and reserves \citep{charles-2001}. Together, these studies establish that geological structure should be represented through multiple plausible interpretations rather than treated as uniquely known.

These ensemble-based methods are typically conditioned on observations from a specific site and generate alternatives to a single interpreted structural model. The resulting ensembles are therefore often limited in size or structural diversity compared with the large and varied populations of geometries required to train ML surrogates. Procedural generation offers an alternative, producing geological diversity systematically rather than by perturbing an interpreted model. \citet{jessell-2022} introduced the Noddyverse, a large dataset of three-dimensional geological models generated by randomising and recombining kinematic geological process simulations for ML and inversion applications. \citet{ghyselincks-2025} adopted a similar approach, sampling folding, faulting, and magmatic intrusion within a stochastic framework to generate diverse three-dimensional subsurface structures. The in-house structural generator developed here follows the same principle but targets two-dimensional geometries and pairs each geometry with multiphase flow simulations for GCS surrogate modelling.

The simulations follow the physical formulation and domain dimensions of Case B of the 11th Society of Petroleum Engineers Comparative Solution Project (SPE11B) \citep{nordbotten-2024}, which builds on the FluidFlower validation benchmark study \citep{fluidflower-2024}. In FluidFlower, forecasts of CO$_2$ migration were assessed against direct observations in a physically constructed storage formation comprising layered sand facies with contrasting grain sizes and permeabilities, including fine-sand barriers that act as local seals \citep{fluidflower-2024}. SPE11B scales this configuration to reservoir dimensions and introduces geothermal boundary conditions and thermal effects absent from the original experiment. This design emphasises structural heterogeneity, including facies layering and faults, rather than fine-scale petrophysical variability, retaining geological complexity while keeping the simulation problem tractable. The formulation includes structural and solubility trapping, with residual trapping represented through an immobile gas saturation. Unlike SPE11B, which prescribes a single reference geometry and two fixed injection wells, \textit{FlowersBench} samples structural geometries and assigns one stochastically placed injection well to each simulation case.

\section{Methods} \label{sec:methodology}

\begin{figure}[htbp]
    \centering
    \includegraphics[width=0.8\textwidth]{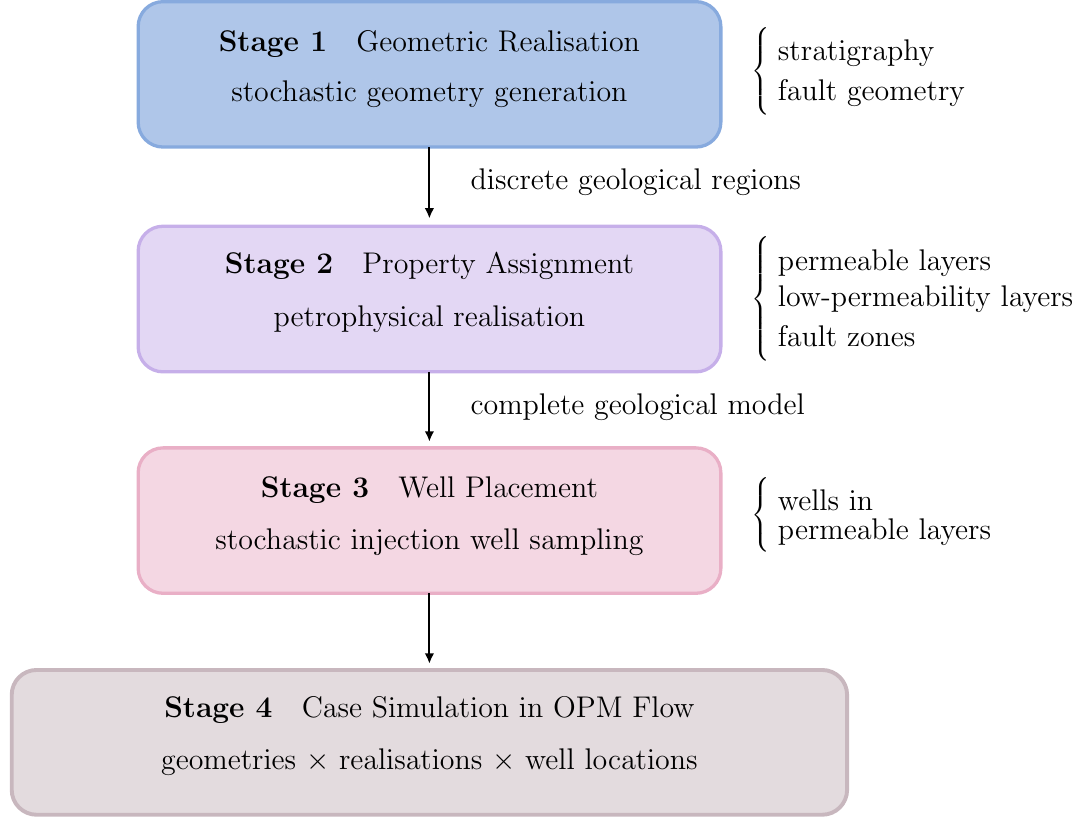}
    \caption{Four-stage workflow used to generate \textit{FlowersBench}: geometric realisation, property assignment, well placement, and case simulation in OPM Flow.}
    \label{fig:four_stage_workflow}
\end{figure}

\textit{FlowersBench} is generated through a four-stage workflow, as shown in Figure \ref{fig:four_stage_workflow}. First, the in-house structural generator samples structural geometries featuring labelled stratigraphic layers, fault damage zones, and fault cores. Petrophysical properties are assigned to each region, with layers classified as permeable or low-permeability and fault cores as sealing or non-sealing. Injection wells are then placed within admissible areas of the permeable layers. Each combination of structural geometry, property realisation, and well location defines a case for numerical modelling with OPM Flow (release 2025.04) \citep{rasmussen-2020}. Validity checks are performed at each stage, and configurations that fail these checks are discarded.

The workflow generates multiple simulation cases for each structural geometry, with up to five property realisations per geometry and up to five well locations per realisation. The following sections describe each stage of the workflow: the structural generator (Section \ref{sec:geometry_generation}), petrophysical property assignment (Section \ref{sec:property_sampling}), stochastic well placement (Section \ref{sec:stochastic_well_placement}), and the OPM Flow simulation setup (Section \ref{sec:simulation_setup}).

The released dataset includes 244 structural geometries and 5,000 simulation cases. Each structural geometry is provided as a \texttt{.geo} file of region polygons and a \texttt{.json} file of region labels. Each simulation case is provided as a compressed NumPy archive. The archive stores the static property fields and four dynamic fields: gas saturation, pressure, temperature, and dissolved CO$_2$. The dynamic fields are reported at 17 times between 10 and 550 years after the start of injection. A companion file stores the pressure and temperature fields at $t = 0$, and a property file for each property realisation stores the
immobile wetting saturation, capillary entry pressure, and dispersivity. The file structure and stored variables are described in Appendix \ref{app:data_records}.

\subsection{Stochastic Geometry Generation} \label{sec:geometry_generation}

Structural geometries are generated on a rectangular domain
\[
\Omega = [x_{\min}, x_{\max}] \times [z_{\min}, z_{\max}] = [0, 2.8] \times [0, 1.2],
\]
expressed in normalised, dimensionless coordinates. During simulation setup, this domain is rescaled anisotropically to the physical dimensions of SPE11B, 8.40 km $\times$ 1.20 km (Section \ref{sec:simulation_setup}).

Each structural geometry $X$ is sampled from a stochastic generator $G$,
\begin{equation}
X = G(Z,\,\Theta).
\end{equation}
Here, $Z$ encodes discrete structural choices, including the number of stratigraphic layers and the fault configuration. $\Theta$ specifies continuous geometric parameters that control the shape and position of horizons and the location, dip, and throw of faults. Together, $Z$ and $\Theta$ define the structural variability represented in \textit{FlowersBench}. Their sampling distributions are given in Appendix \ref{app:generator-params}.

Stratigraphy is represented by sinusoidal horizons, each spanning the full domain width and dividing an existing region into two layers. A horizon $h_k(x)$ is defined as
\begin{equation}
h_k(x) = z_{\mathrm{bottom}}(x) + h_{\mathrm{region}} \bigl[\, b_k + a_k \sin\bigl(2\pi f_k \hat{x} + \psi_k\bigr) \bigr],
\end{equation}
where $z_{\mathrm{bottom}}(x)$ is the elevation of the lower boundary of the region being divided, $h_{\mathrm{region}}$ is the vertical extent of that region, $b_k \in (0,1)$ is the base height fraction, $a_k$ is the amplitude fraction, $f_k$ is the number of sinusoidal cycles across the domain width, $\psi_k$ is the phase shift, and $\hat{x} = x / 2.8$ is the normalised horizontal coordinate. Each horizon therefore follows the shape of the boundary beneath it, with a sinusoidal perturbation superimposed. The resulting horizons define both the faulted sequence and a set of unfaulted layers within the basal unit beneath it. Examples of structural geometries with varying stratigraphic complexity are shown in Figure \ref{fig:strat_variability}, and the sampling distributions of all horizon parameters are listed in Appendix \ref{app:generator-params}.

\begin{figure}[!htbp]
    \centering
    \includegraphics[width=1.0\textwidth]{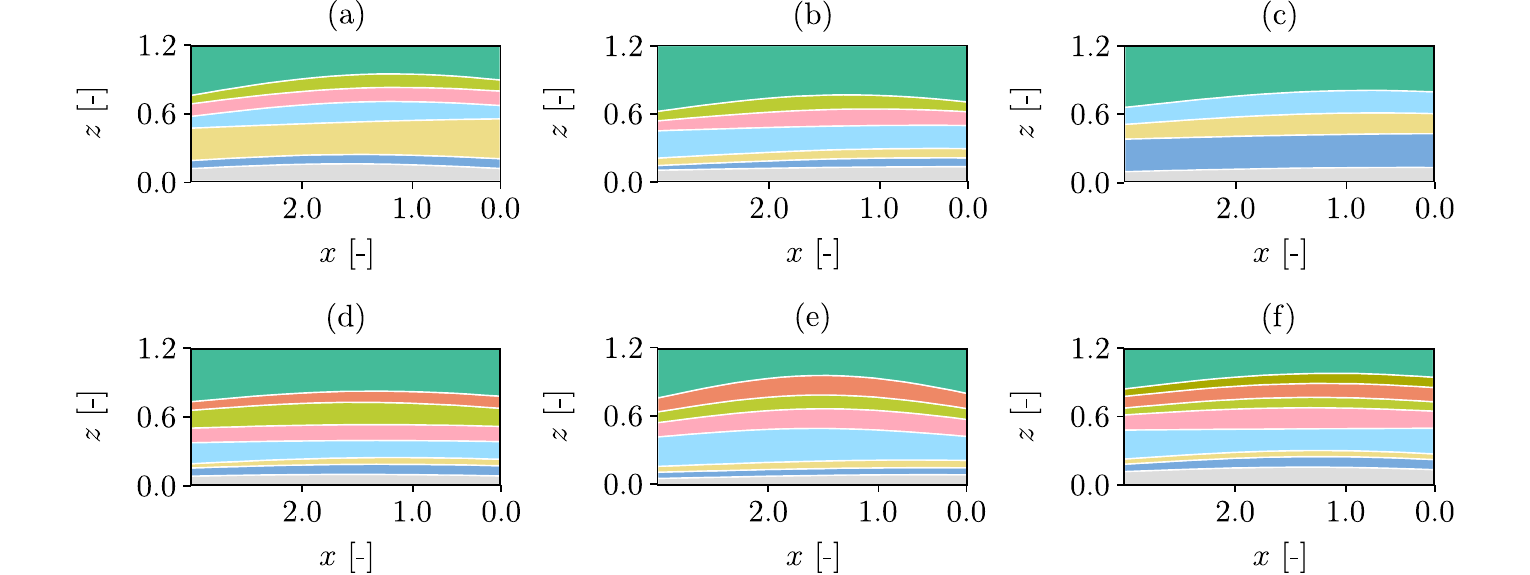}
    \caption{Examples of stratigraphic layering sampled by the in-house structural generator, shown without faults. Colours denote distinct stratigraphic layers. Coordinates are normalised generator coordinates, with $x = 0$ at the right of each panel and $z$ the normalised elevation.}
    \label{fig:strat_variability}
\end{figure}

Fault zones are then introduced into the layered stratigraphy. The fault geometry is defined by a horizontal centre $x_{\mathrm{loc}}$, a fault-pair width $w$, a tip elevation $z_{\mathrm{top}}$, a throw $s$, and a dip angle $\gamma$ for each fault. The centre is restricted to the interior of the domain to avoid boundary artefacts. Dip angles are sampled independently for each fault between $55^\circ$ and $65^\circ$, typical of normal faults \citep{fossen-2016}, and converted to fault-plane slopes in normalised coordinates such that the sampled dip is preserved after rescaling to the physical domain. The generator constructs faults geometrically and does not enforce fault kinematics, so the sense of displacement is not constrained to normal faulting.

Faults are represented as zones of finite thickness. Each fault zone includes a damage zone enclosing a fault core, with fixed thicknesses of $0.050$ and $0.015$ in normalised coordinates, respectively, allowing the two to receive distinct petrophysical properties \citep{fossen-2016}.

\begin{figure}[htbp]
    \centering
    \includegraphics[width=1.0\textwidth]{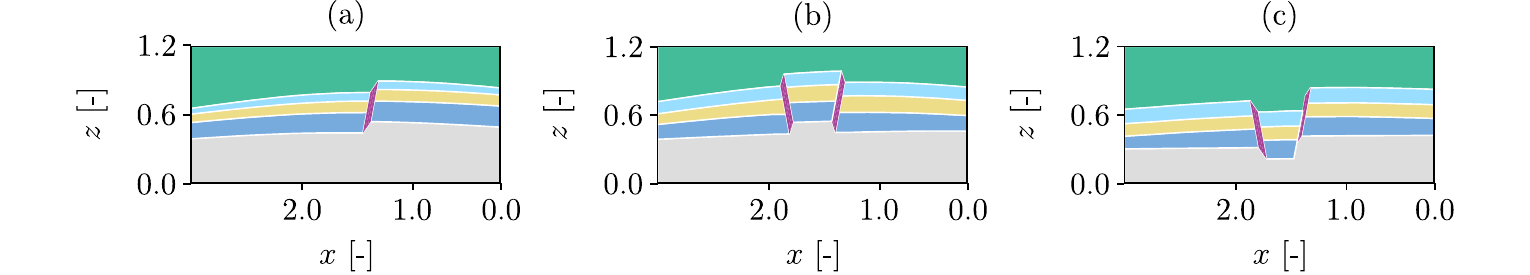}
    \caption{Fault styles sampled by the in-house structural generator: (a) a single fault, (b) a fault pair with a raised central block, and (c) a fault pair with a lowered central block (graben). Fault damage zones are shown in purple and fault cores in a darker shade. Coordinates as in Figure \ref{fig:strat_variability}.}
    \label{fig:fault_styles} 
\end{figure}

Three fault styles are considered: a single fault, and a pair of faults dipping towards each other, with the central block either raised or lowered relative to the adjacent blocks. These styles are sampled with probabilities $0.30$, $0.45$, and $0.25$, respectively. The configuration with a lowered central block corresponds to a graben, a common extensional structure in sedimentary basins \citep{fossen-2016}. The configuration with a raised central block resembles an inverted graben, in which steep faults formed under extension are later reactivated with reverse displacement. In the single-fault style, one fault is placed at $x_{\mathrm{loc}} + w/2$, and the downthrown side is sampled at random. Both faults of a pair share the same throw. The three fault styles are illustrated in Figure \ref{fig:fault_styles}.

For the paired styles, a supplementary fault is inserted with probability $0.30$, positioned away from the fault pair, with a smaller throw and a lower tip elevation. The dip direction and downthrown side of this fault are sampled at random, and its dip of $25^\circ$--$36^\circ$ is shallower than that of the main faults. 

Following fault insertion, the domain is partitioned into labelled regions comprising stratigraphic layers, fault damage zones, and fault cores. Each fault zone is indexed by the order in which the fault was inserted, and these labels are used during petrophysical property assignment (Section \ref{sec:property_sampling}).

Geometric validity is checked after each horizon and fault insertion. Every region is tested for self-intersection and invalid boundaries using \texttt{Shapely}, and geometries that fail these tests are rejected and regenerated. Each retained geometry is exported as a Gmsh \citep{geuzaine-2009} geometry (\texttt{.geo}) file that stores the polygon boundaries of all regions, for use in petrophysical property assignment and simulation grid construction (Sections \ref{sec:property_sampling} and \ref{sec:simulation_setup}).

\subsection{Petrophysical Property Assignment} \label{sec:property_sampling}

Each region of the structural geometry is first assigned to a property class, and petrophysical properties are then sampled from populations specific to that class. Stratigraphic layers are classified as permeable or low-permeability, and fault cores as sealing or non-sealing, following the scheme summarised in Algorithm \ref{alg:facies_classification}. Two layer positions are fixed to represent a confined storage system. The lowermost layer forms an impermeable basal layer that prevents downward migration of CO$_2$, and the topmost layer forms a low-permeability caprock that limits upward migration, consistent with the SPE11B benchmark \citep{nordbotten-2024}.

\begin{algorithm}[htbp]
\caption{Classification of stratigraphic layers and fault zones.}
\label{alg:facies_classification}
\DontPrintSemicolon
\SetAlgoLined
\SetInd{0.4em}{1.0em}
\BlankLine
\textbf{Stratigraphic layers}\;
\Indp
Lowermost layer $\rightarrow$ impermeable basal layer\;
Topmost layer $\rightarrow$ low-permeability caprock\;
Compute the fractional area $a_i$ of each intermediate layer $i$\;
Build the candidate set $\mathcal{C}$:\;
\Indp
add layers with $a_i < f_{\rm res}$\;
add layers with $a_i \geq f_{\rm res}$ and area below the group median\;
add remaining layers with $a_i \geq f_{\rm res}$ only if part of the faulted sequence\;
remove layers adjacent to the lowermost or topmost layer\;
\Indm
Draw one layer from $\mathcal{C}$ $\rightarrow$ low-permeability\;
All remaining layers $\rightarrow$ permeable\;
\Indm
\BlankLine
\textbf{Fault zones}\;
\Indp
Fault damage zones $\rightarrow$ fault-specific populations\;
Fault cores:\;
\Indp
low-permeability layer in faulted sequence $\rightarrow$ sealing with probability $p_{\rm sealing}$\;
otherwise $\rightarrow$ non-sealing\;
\Indm
\Indm
\BlankLine
\end{algorithm}

In addition to the two fixed layers, one intermediate layer is assigned as low-permeability. Each intermediate layer $i$ is assigned a fractional area $a_i$ relative to the total domain area. Layers with $a_i < f_{\rm res} = 0.07$ represent thin stratigraphic units, such as shale or mudstone interbeds, and are added to a candidate set $\mathcal{C}$. Layers with $a_i \geq f_{\rm res}$ are added to $\mathcal{C}$ if their area is below the median area of this group, and otherwise only if part of the faulted sequence. Layers immediately adjacent to the lowermost or topmost layer are removed from $\mathcal{C}$ to avoid consecutive low-permeability layers.

The low-permeability layer is then drawn from $\mathcal{C}$, with the candidates taken in random order and reused only when a geometry has more property realisations than candidates. All remaining layers are classified as permeable. Geometries with only one layer above $f_{\rm res}$, or with no remaining candidates in $\mathcal{C}$, are discarded.

Fault cores are classified as sealing or non-sealing. If the additional low-permeability layer is part of the faulted sequence, each fault core is designated as sealing with probability $p_{\rm sealing} = 0.40$. Otherwise, all fault cores are non-sealing. This reflects the entrainment of low-permeability material into fault cores during faulting \citep{salo-salgado-2023}.

Once classified, each region is assigned a facies, which defines the population from which its properties are drawn. Some properties are fixed across all facies, while others are sampled. The fixed properties are the immobile non-wetting saturation $S_{n,\rm imm}$, the maximum capillary pressure $p_{\rm cap,max}$, the dispersivity $E$, the rock specific heat capacity $c_s$, and the rock density $\rho_s$. These follow the SPE11B benchmark \citep{nordbotten-2024} and are listed in Table \ref{tab:fixed_properties}.

\begin{table}[htbp]
\centering
\caption{Fixed petrophysical properties, following the SPE11B benchmark \citep{nordbotten-2024}. Values apply to all facies except the impermeable basal layer, for which $S_{n,\rm imm}$ and $E$ are set to zero.}
\label{tab:fixed_properties}
\renewcommand{\arraystretch}{1.4}
\begin{tabular}{l@{\hspace{4em}}l}
\toprule
Immobile non-wetting saturation, $S_{n,\rm imm}$ (-) & 0.1 \\
Maximum capillary pressure, $p_{\rm cap,max}$ (Pa) & $3 \times 10^7$ \\
Dispersivity, $E$ (m) & 10 \\
Rock specific heat capacity, $c_s$ (kJ\,kg$^{-1}$\,K$^{-1}$) & 0.85 \\
Rock density, $\rho_s$ (kg\,m$^{-3}$) & 2500 \\
\bottomrule
\end{tabular}
\end{table}

All other properties are sampled from facies-specific populations. For each facies, a population of $N = 1{,}000$ property sets is generated in advance. Each property is drawn independently from a Gaussian distribution centred on a facies-specific mean $\mu$, with a standard deviation of $0.15\mu$, and clipped to within $\pm 25\%$ of the mean and to class-specific bounds,
\begin{equation}
    \tilde{x} \sim \mathcal{N}\!\left(\mu,\, (0.15\,\mu)^2\right),
    \qquad
    \tilde{x} \leftarrow \mathrm{clip}\!\left(\tilde{x},\, \max(0.75\mu,\, x_{\rm low}),\, \min(1.25\mu,\, x_{\rm high})\right),
\end{equation}
where $x_{\rm low}$ and $x_{\rm high}$ are the lower and upper bounds of the corresponding class. The sampled properties are porosity $\phi$, horizontal intrinsic permeability $K$, immobile wetting saturation $S_{w,\rm imm}$, and rock thermal conductivity $\kappa_s$. For each region, one property set is drawn at random from the population of its facies. The facies-specific means are listed in Tables \ref{tab:facies_means_layers} and \ref{tab:facies_means_faults}.

The capillary entry pressure $p_{\rm entry}$ is not sampled independently but is derived from the sampled porosity and permeability using Leverett $J$-scaling,
\begin{equation}
    p_{\rm entry} = \sigma_J \sqrt{\frac{\phi}{K}}, \qquad \sigma_J = 6.12 \times 10^{-3}\ \mathrm{N\,m^{-1}},
\end{equation}
with $K$ in m$^2$. The scaling constant follows \citet{abdoulghafour-2019}, consistent with the SPE11B benchmark \citep{nordbotten-2024}.

Permeable facies are labelled P1--P7 and low-permeability facies L1--L3. Within each class, facies are assigned to layers in order of vertical position, from the bottom of the sequence upward. Any permeable layers above the seventh are assigned P7, which occurs only in the parallel-fault subset (Appendix~\ref{app:shift_subsets}). Permeability in the permeable layers therefore increases towards the top of the sequence, representing a coarsening-upward trend associated with increasing depositional energy. This ordering is a modelling assumption and does not reproduce the facies arrangement of SPE11B. The permeable facies are adapted from SPE11B. P1 uses the properties of SPE11B facies 1, which has the lowest non-zero permeability in SPE11B, and is the least permeable reservoir facies here. Low-permeability layers are represented by facies L1 and L2, assigned to the intermediate layer and the caprock, respectively; L3 is used only for sealing fault cores. The lowermost layer is assigned fixed near-zero properties, representing an impermeable basal layer with $K = 10^{-5}$\,mD, $\phi = 10^{-6}$, $S_{w,\rm imm} = 0$, $\kappa_s = 2.0$\,W\,m$^{-1}$\,K$^{-1}$, and $p_{\rm entry} = 0$\,Pa.

\begin{table}[htbp]
\centering
\caption{Population means for the stratigraphic facies. Facies are assigned in order from the bottom of the sequence upward, and sampled values are clipped to the bounds listed for each class.}
\label{tab:facies_means_layers}
\renewcommand{\arraystretch}{1.15}
\begin{tabular}{lcccc}
\toprule
Facies & $\phi$ (-) & $K$ (mD) & $S_{w,\rm imm}$ (-) & $\kappa_s$ (W\,m$^{-1}$\,K$^{-1}$) \\
\midrule
\multicolumn{5}{l}{\textit{Permeable}} \\
P1 & 0.10 & 0.10   & 0.32 & 1.90 \\
P2 & 0.20 & 101.3  & 0.14 & 1.25 \\
P3 & 0.20 & 202.7  & 0.12 & 1.25 \\
P4 & 0.20 & 506.6  & 0.12 & 1.25 \\
P5 & 0.22 & 709.3  & 0.12 & 1.10 \\
P6 & 0.24 & 861.0  & 0.12 & 1.00 \\
P7 & 0.25 & 1013.3 & 0.12 & 0.92 \\
Bounds & [0.08, 0.28] & [0.05, 1500] & [0.08, 0.32] & [0.5, 2.5] \\
\midrule
\multicolumn{5}{l}{\textit{Low-permeability}} \\
L1 & 0.060 & 0.0010 & 0.35 & 1.80 \\
L2 & 0.065 & 0.0015 & 0.34 & 1.775 \\
L3 & 0.070 & 0.0020 & 0.33 & 1.75 \\
Bounds & [0.02, 0.12] & [$10^{-4}$, 0.01] & [0.20, 0.50] & [1.0, 2.5] \\
\bottomrule
\end{tabular}
\end{table}

Fault damage zones are assigned properties from two damage-zone populations, indexed by the order in which the faults were inserted. Non-sealing fault cores are assigned properties from two fault-core populations, indexed in the same way, and the supplementary fault uses the second population of each type (Table \ref{tab:facies_means_faults}). Sealing fault cores instead use the low-permeability populations in Table \ref{tab:facies_means_layers}, with L1, L2, and L3 assigned to the first, second, and supplementary faults, respectively.

\begin{table}[htbp]
\centering
\caption{Population means for the fault facies. Populations are indexed by the order in which the faults were inserted, and sampled values are clipped to the bounds listed.}
\label{tab:facies_means_faults}
\renewcommand{\arraystretch}{1.15}
\begin{tabular}{lcccc}
\toprule
Facies & $\phi$ (-) & $K$ (mD) & $S_{w,\rm imm}$ (-) & $\kappa_s$ (W\,m$^{-1}$\,K$^{-1}$) \\
\midrule
\multicolumn{5}{l}{\textit{Damage zone}} \\
D1 & 0.34 & 1800 & 0.11 & 0.26 \\
D2 & 0.32 & 1500 & 0.12 & 0.28 \\
\midrule
\multicolumn{5}{l}{\textit{Non-sealing fault core}} \\
C1 & 0.32 & 1000 & 0.12 & 0.26 \\
C2 & 0.30 & 800  & 0.13 & 0.28 \\
\midrule
Bounds & [0.20, 0.40] & [100, 3000] & [0.05, 0.20] & [0.20, 1.0] \\
\bottomrule
\end{tabular}
\end{table}

\subsection{Stochastic Well Placement} \label{sec:stochastic_well_placement}

Each property realisation is paired with up to five injection-well locations, each defining a separate simulation case. Well locations are sampled within the permeable layers, subject to constraints on layer thickness, distance from faults and domain boundaries, and proximity to low-permeability layers.

Candidate layers for well placement are the stratigraphic layers with a permeability of at least $K_{\min}$ and a vertical extent of at least $t_{\min}$,
\begin{equation}
    \mathcal{L} = \left\{ l \notin \mathcal{L}_{\rm fault} \mid K_l \geq K_{\min},\ t_l \geq t_{\min} \right\},
\end{equation}
where $\mathcal{L}_{\rm fault}$ denotes the set of fault zones. Layer boundaries are sinusoidal and offset by faults, and layer thickness therefore varies laterally. The local thickness is also evaluated at each sampled horizontal position, and positions where it falls below $t_{\min}$ are rejected.

A layer is first selected from $\mathcal{L}$ with probability inversely proportional to the elevation of its centroid, so that deeper layers are preferred. A horizontal position $x$ is then sampled uniformly within the selected layer, after shrinking the layer polygon inward by a buffer $\delta_l$. Positions within $\delta_{\rm domain}$ of the lateral domain boundaries, or within $\delta_f$ of any fault zone, are rejected,
\begin{equation}
x \in [\delta_{\rm domain},\, x_{\max} - \delta_{\rm domain}],
\qquad
x \notin \bigcup_k \bigl[x_k^{-} - \delta_f,\, x_k^{+} + \delta_f\bigr],
\end{equation}
where $x_k^{-}$ and $x_k^{+}$ are the horizontal limits of fault zone $k$.

The vertical position is sampled within the shrunken layer as $z = z_{\min} + u\,(z_{\max} - z_{\min})$, with $u \sim \mathrm{Beta}(0.5, 5)$, favouring the lower part of the layer. Locations within $\delta_z$ of the vertical extent of any low-permeability layer are rejected, and a minimum distance $\delta_w$ is enforced between wells sampled for the same property realisation. Rejected locations are resampled. If no valid location is found, $\delta_w$ and $\delta_z$ are relaxed stepwise to half their values and then to zero. All placement parameters are listed in Table \ref{tab:well_placement_params}.

\begin{table}[htbp]
\centering
\caption{Parameters for stochastic well placement, in normalised generator coordinates and physical units. Horizontal and vertical lengths scale by 3000 and 1000, respectively, so $\delta_l$ and $\delta_w$, which act in both directions, span a range of physical values.}
\label{tab:well_placement_params}
\renewcommand{\arraystretch}{1.3}
\begin{tabular}{lcc}
\toprule
Parameter & Normalised & Physical \\
\midrule
Minimum permeability, $K_{\min}$ & - & 0.05\,mD \\
Minimum layer thickness, $t_{\min}$ & 0.15 & 150\,m \\
Layer boundary buffer, $\delta_l$ & 0.05 & 50--150\,m \\
Domain boundary buffer, $\delta_{\rm domain}$ & 0.30 & 900\,m \\
Fault exclusion buffer, $\delta_f$ & 0.05 & 150\,m \\
Low-permeability layer buffer, $\delta_z$ & 0.05 & 50\,m \\
Minimum inter-well distance, $\delta_w$ & 0.10 & 100--300\,m \\
Beta shape parameters, $(\alpha, \beta)$ & (0.5, 5.0) & - \\
\bottomrule
\end{tabular}
\end{table}

\subsection{OPM Reservoir Simulation} \label{sec:simulation_setup}

Each simulation case is run with OPM Flow \citep{rasmussen-2020}, a fully implicit finite-volume reservoir simulator, following the SPE11B specification \citep{nordbotten-2024}. Simulation decks are generated with a modified version of \texttt{pyopmspe11} \citep{landa-marban-2025}, a Python framework for setting up and running the SPE11 cases in OPM Flow. Only the structural geometry, petrophysical properties, and well location vary between cases. The grid, fluid model, initialisation settings, boundary conditions, and simulation schedule are identical for all cases.

The normalised generator domain is rescaled to the physical simulation domain,
\[
\Omega = [0, 2.8] \times [0, 1.2]
\qquad \longrightarrow \qquad
\Omega_{\mathrm{phys}} = [0, 8400] \times [0, 1200]\ \mathrm{m},
\]
using scale factors of 3000 in $x$ and 1000 in $z$. This rescaling is anisotropic and does not preserve angles, which is accounted for when sampling fault dips (Section \ref{sec:geometry_generation}). A unit thickness of 1\,m is assumed in the out-of-plane direction. The physical domain is discretised on a Cartesian grid of $842 \times 120$ cells. All cells are 10\,m high and interior cells are 10\,m wide, while the outermost column on each side is split into cells of 1\,m and 9\,m, with the 1\,m cells forming the lateral boundary.

Each grid cell is assigned to a region by a point-in-polygon test of its centre against the region polygons stored in the \texttt{.geo} file (Section \ref{sec:geometry_generation}). Per-cell fields of horizontal permeability (\texttt{PERMX}), porosity, rock thermal conductivity, and dispersivity are then set from the property set of its region (Section \ref{sec:property_sampling}), with vertical permeability $\texttt{PERMZ} = 0.1\,\texttt{PERMX}$. Each region is also given its own saturation table (\texttt{SATNUM}), which defines its relative permeability and capillary pressure functions following the SPE11B functional forms \citep{nordbotten-2024}.

Flow, transport, and thermal behaviour are modelled with the \texttt{CO2STORE} module of OPM Flow, representing a two-phase system of brine and a CO$_2$-rich gas phase (governing equations in Appendix \ref{app:governing_equations}). CO$_2$ dissolves in the brine (\texttt{DISGASW}) and water vaporises into the gas phase (\texttt{VAPWAT}). Molecular diffusion (\texttt{DIFFUSE}) uses uniform coefficients of $10^{-9}$\,m$^2$\,s$^{-1}$ in the liquid phase and $2 \times 10^{-8}$\,m$^2$\,s$^{-1}$ in the gas phase, and mechanical dispersion uses the per-cell dispersivity. Thermal effects are included (\texttt{THERMAL}), with a uniform rock heat capacity and density (Table \ref{tab:fixed_properties}).

The initial state is hydrostatic, with a pressure of 300\,bar at a depth of 900\,m below the top of the domain (\texttt{EQUIL}). The initial temperature increases linearly from 40\,$^\circ$C at the top to 70\,$^\circ$C at the bottom (\texttt{RTEMPVD}), consistent with the geothermal gradient of 25\,$^\circ$C\,km$^{-1}$ in SPE11B. The top and bottom boundaries are closed to flow and held at these temperatures. The lateral boundary cells are assigned large pore volumes, emulating the embedding of the domain in a larger aquifer, following SPE11B \citep{nordbotten-2024}.

Injection is represented by a CO$_2$ mass source (\texttt{SOURCE}) in the grid cell whose centre is nearest to the sampled well location (Section \ref{sec:stochastic_well_placement}). CO$_2$ is injected at 0.035\,kg\,s$^{-1}$ and 10\,$^\circ$C, matching the rate of a single SPE11B well. The simulation input retains the second well of the SPE11B template, so that all decks share the standard template structure, but its source rate is zero throughout. A zero-rate source adds neither mass nor energy, and each case therefore has a single active injection well.

Each simulation starts with a 999.9-year equilibration period, followed by a 0.1-year settling period, both without injection. Together, these correspond to the 1000-year initialisation period of SPE11B, and the state at the end of the settling period defines the initial condition ($t = 0$) of each case. CO$_2$ is then injected at 10\,$^\circ$C for 50 years, followed by a 500-year post-injection monitoring period (Table \ref{tab:injection_schedule}). Results are reported every 5 years from the start of injection.

\begin{table}[htbp]
\centering
\caption{Simulation schedule, identical for all cases.}
\label{tab:injection_schedule}
\renewcommand{\arraystretch}{1.25}
\begin{tabular}{lccc}
\toprule
Period & Duration, $\Delta t$ (yr) & Max.\ time step (yr) & Injection rate, $q$ (kg\,s$^{-1}$) \\
\midrule
Equilibration             & 999.9 & 1.0 & 0     \\
Settling                  & 0.1   & 0.1 & 0     \\
Injection                 & 50    & 0.1 & 0.035 \\
Post-injection monitoring & 500   & 0.1 & 0     \\
\bottomrule
\end{tabular}
\end{table}

Each case is solved with the fully implicit Newton solver of OPM Flow, with adaptive time stepping limited by the maximum step sizes in Table \ref{tab:injection_schedule}. Simulations are run in parallel using MPI domain decomposition. The released fields (Appendix~\ref{app:data_records}) are extracted from the restart files written at each report step.

\section{Dataset Characterisation} \label{sec:dataset_characterisation}

\textit{FlowersBench} is designed to expose surrogates to variability in geological structure, as well as in rock properties and well locations. This section characterises this variability, first in the sampled geology and well locations, and then in the resulting storage behaviour. Variability arises at two levels: structural geometry determines the stratigraphy and fault configuration, while property realisations and well locations vary across cases sharing the same geometry. All analyses use the 5,000 simulation cases in the main dataset, excluding reflected samples.

\begin{figure}[htbp]
    \centering
    \includegraphics[width=1.0\textwidth]{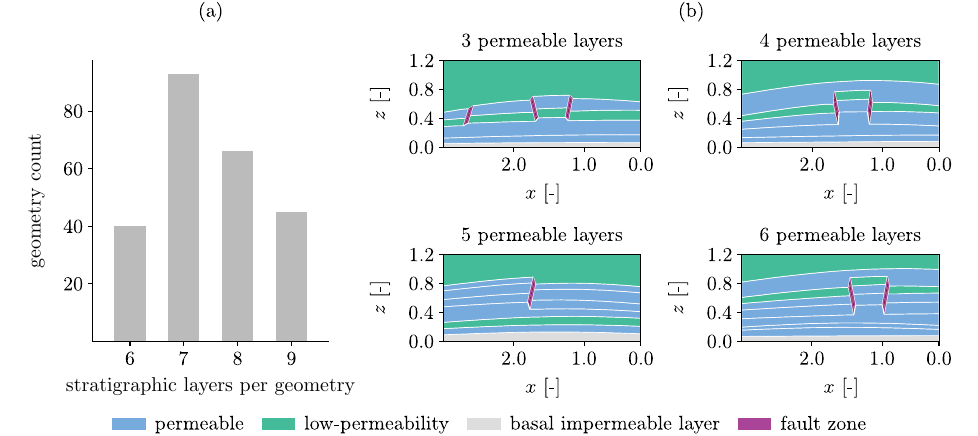}
    \caption{Stratigraphic layering across the 244 structural geometries: (a) number of stratigraphic layers per geometry, and (b) example geometries with three to six permeable layers. Coordinates as in Figure \ref{fig:strat_variability}.}
    \label{fig:structural_diversity}
\end{figure}

Table \ref{tab:static_summary_stats} summarises the sampled structure and the permeability of each facies class.

\begin{table}[htbp]
\centering
\caption{Summary statistics of the sampled structure and permeability. $N$ is the number of geometries, property realisations, or regions summarised in each row.}
\label{tab:static_summary_stats}
\renewcommand{\arraystretch}{1.2}
\begin{tabular}{lrrrrr}
\toprule
 & $N$ & Mean & Median & Min & Max \\
\midrule
Stratigraphic layers per geometry & 244 & 7.48 & 7 & 6 & 9 \\
Permeable layers per property realisation & 1,146 & 4.48 & 4 & 3 & 6 \\
\midrule
\multicolumn{6}{l}{\textit{Permeability (mD)}} \\
Permeable layers & 5,139 & 271.6 & 180.0 & 0.076 & 1076.0 \\
Low-permeability layers & 2,292 & 0.0013 & 0.0012 & 0.0008 & 0.0019 \\
Sealing fault cores & 576 & 0.0012 & 0.0011 & 0.0008 & 0.0025 \\
Fault damage zones & 1,875 & 1687.9 & 1679.0 & 1125.0 & 2250.0 \\
Non-sealing fault cores & 1,299 & 928.3 & 915.2 & 600.0 & 1250.0 \\
\bottomrule
\end{tabular}
\end{table}

Stratigraphy varies considerably across geometries. The number of stratigraphic layers ranges from 6 to 9, with 7 layers being the most common (Figure \ref{fig:structural_diversity} (a)). All property realisations contain three layers that are not permeable: the impermeable basal layer, the caprock, and an intermediate low-permeability layer. The remaining 3 to 6 layers therefore constitute the permeable part of the sequence (Figure \ref{fig:structural_diversity} (b)). The geometries are represented relatively evenly in the dataset, with most contributing between 19 and 24 simulation cases. Of the 244 geometries, 211 have five property realisations, and the dataset contains 1,146 property realisations in total.

Fault configurations introduce a second source of structural variability. Of the 244 geometries, 43.0\% have a single fault, 48.4\% a fault pair, and 8.6\% a fault pair with a supplementary fault. Including those with a supplementary fault, fault pairs with a raised central block are approximately three times as common as those with a lowered central block, accounting for 42.6\% and 14.3\% of all geometries, respectively. These proportions differ from the sampling probabilities in Section \ref{sec:geometry_generation}, reflecting geometries rejected during generation and at subsequent stages of the workflow.

\begin{figure}[htbp]
    \centering
    \includegraphics[width=\textwidth]{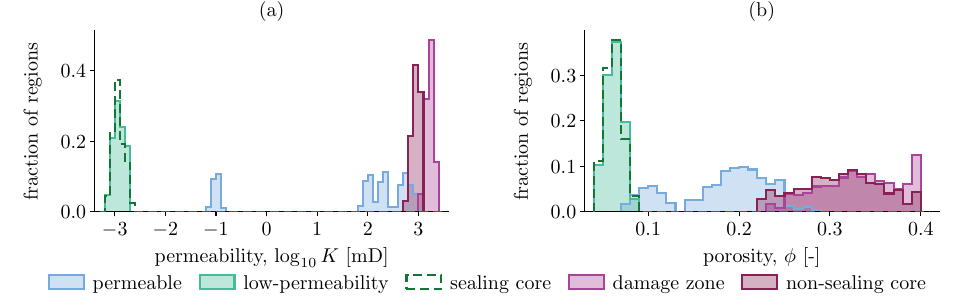}
    \caption{Distributions of (a) permeability, on a logarithmic scale, and (b) porosity by facies class across the 1,146 property realisations. Each region contributes one value, and each histogram shows the fraction of regions of that class in each bin. Sealing fault cores are shown as a dashed outline.}
    \label{fig:facies_property_densities}
\end{figure}

Property realisations show pronounced contrasts between facies classes (Figure~\ref{fig:facies_property_densities}). Permeable layers have a mean permeability of 271.6\,mD, approximately five orders of magnitude higher than that of low-permeability layers, which act as barriers to vertical migration. Fault damage zones and non-sealing fault cores have higher mean permeabilities of 1687.9\,mD and 928.3\,mD, respectively, and may allow CO$_2$ to bypass these barriers. In contrast, sealing cores, which account for 30.7\% of all fault cores, restrict flow across faults.

\begin{figure}[htbp]
    \centering
    \includegraphics[width=0.7\textwidth, height=6cm, keepaspectratio]{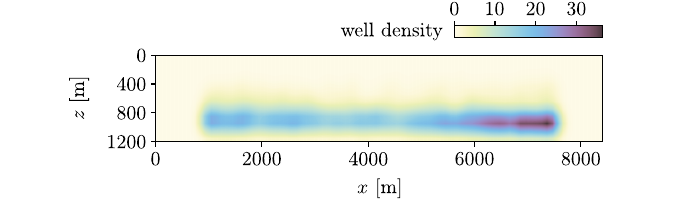}
    \caption{Density of the 5,000 sampled injection-well locations. Coordinates are physical, with $x$ measured from the left boundary of the domain and $z$ increasing downward from the top.}
    \label{fig:well_placement_coverage}
\end{figure}

Well locations introduce further variability among cases that share the same geology (Figure \ref{fig:well_placement_coverage}). 

Wells span the domain between the 900\,m lateral buffers but are concentrated at depth, with half located between 835 and 945\,m. This distribution reflects the preference for deeper layers and for lower positions within each layer. Well density is also highest between approximately 6,000 and 7,500\,m, which lies largely outside the range of main fault locations.

\begin{figure}[htbp]
    \centering
    \includegraphics[width=1.0\textwidth, height=5cm, keepaspectratio]{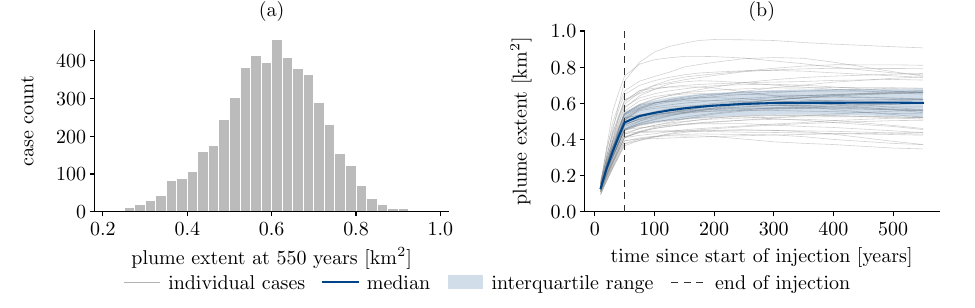}
    \caption{CO$_2$ plume extent across the 5,000 simulation cases, defined by $S_g > 0.01$: (a) distribution at 550 years, and (b) median over time, with the interquartile range and 50 randomly selected individual cases. The dashed line marks the end of injection.}
    \label{fig:dynamic_outcome_diversity_and_plume_extent}
\end{figure}

These differences in geology and well location translate into a wide range of storage behaviour. A simple measure of this behaviour is plume extent, the cross-sectional area of cells with a gas saturation above 0.01. At 550 years, plume extent ranges from 0.22 to 0.96\,km$^2$, with most cases close to the mean of 0.60\,km$^2$ (Figure \ref{fig:dynamic_outcome_diversity_and_plume_extent} (a)). Plumes grow rapidly during injection, from a median of 0.13\,km$^2$ at 10 years to 0.49\,km$^2$ at 50 years, and then expand more slowly before levelling off at about 0.60\,km$^2$ after 300 years (Figure \ref{fig:dynamic_outcome_diversity_and_plume_extent} (b)). The spread between cases keeps increasing throughout, with the interquartile range widening from 0.03 to 0.16\,km$^2$. Differences between cases therefore continue to develop long after injection ends, as buoyant migration controlled by the geology redistributes the CO$_2$.

\begin{figure}[htbp]
    \centering
    \includegraphics[width=1.0\textwidth]{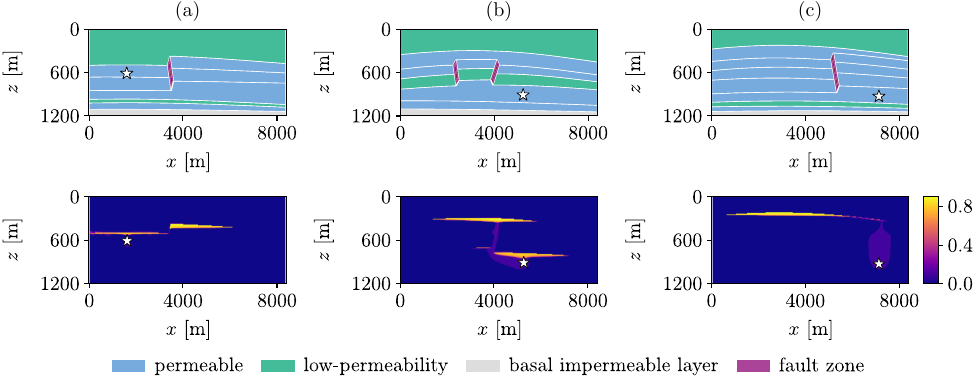}
    \caption{Three example cases, (a)--(c). The top row shows the facies classes of each structural geometry, and the bottom row shows the gas saturation $S_g$ at 550 years. Coordinates as in Figure \ref{fig:well_placement_coverage}. Stars mark the injection wells.}
    \label{fig:representative_case_gallery}
\end{figure}

Plume extent alone does not reveal the differences in CO$_2$ migration pathways. The three cases in Figure \ref{fig:representative_case_gallery} illustrate how these pathways vary across simulations. 

In case (a), CO$_2$ crosses a fault and accumulates in two separate compartments. In case (b), it rises along a fault zone into a shallower layer, leaving a trail of residual gas behind. In case (c), it rises from the well and travels several kilometres laterally beneath the caprock. Cases with similar plume extents can therefore follow very different migration patterns, and the full saturation fields, rather than scalar summaries, are used as surrogate targets.

The geological variability also shapes the pressure, thermal, and dissolution responses (Table \ref{tab:outcome_summary}). The maximum pressure increase relative to the initial state averages 197\,bar across cases and, in 61\% of cases, is reached at the end of injection. Injection of CO$_2$ at 10\,$^\circ$C lowers the reservoir temperature by a maximum of 52\,$^\circ$C on average. By 550 years, the domain-mean dissolved CO$_2$ content averages 3.28\,sm$^3$\,sm$^{-3}$ and varies by a factor of three between cases.

\begin{table}[htbp]
\centering
\caption{Summary statistics of outcome metrics across the 5,000 simulation cases. Changes in pressure and temperature are relative to the initial state at $t = 0$ and are taken over all grid cells and reporting times.}
\label{tab:outcome_summary}
\renewcommand{\arraystretch}{1.2}
\begin{tabular}{llrrrr}
\toprule
Metric & Unit & Mean & Median & Min & Max \\
\midrule
\multicolumn{6}{l}{\textit{Plume extent at 550 years}} \\
\quad $S_g > 0.01$ & km$^2$ & 0.60 & 0.60 & 0.22 & 0.96 \\
\quad $S_g > 0.1$  & km$^2$ & 0.29 & 0.29 & 0.04 & 0.68 \\
\midrule
Maximum pressure increase, $\Delta P_{\max}$ & bar & 196.8 & 178.5 & 83.2 & 648.7 \\
Maximum temperature decrease, $\Delta T_{\max}$ & $^\circ$C & 52.1 & 52.9 & 39.0 & 56.5 \\
Mean dissolved CO$_2$ at 550 years, $R_{\mathrm{sw}}$ & sm$^3$\,sm$^{-3}$ & 3.28 & 3.27 & 1.76 & 5.23 \\
\bottomrule
\end{tabular}
\end{table}

\section{Surrogate Modelling Baseline} \label{sec:surrogate_baseline}

We establish a baseline surrogate on \textit{FlowersBench} to predict the temporal evolution of CO$_2$ gas saturation. The surrogate uses a convolutional encoder-decoder following the U-Net architecture \citep{ronneberger-2015}, with a residual bottleneck. It takes five input fields: porosity, permeability, a well mask encoding the injection location and rate, and the pressure and temperature fields at $t = 0$. The surrogate predicts gas saturation at all 17 reporting times in a single forward pass. This baseline serves as a reference for comparison and is not optimised. The dataset also provides pressure, temperature, and dissolved CO$_2$ as potential surrogate targets. Autoregressive formulations and alternative architectures can be adopted in place of the direct multi-step formulation used here.

The surrogate is evaluated on unseen structural geometries and under structural shift. The first setting uses held-out structural geometries sampled using the same generator configuration as the training data. The second setting uses the two structural-shift subsets, generated with a modified configuration of the structural generator and with a separate parallel-fault generator, respectively. In both settings, we report plume-masked mean absolute error, signed bias, and relative $L_2$ error. Metrics are averaged first within each structural geometry and then across geometries, giving equal weight to each geometry.

\subsection{Problem Formulation} \label{sec:problem_formulation}

Let $x \in \mathbb{R}^{5 \times H \times W}$ denote the five stacked input fields and $y \in \mathbb{R}^{17 \times H \times W}$ the gas saturation at the 17 reporting times, both at the downsampled resolution $H \times W = 64 \times 256$ (Section \ref{sec:data_preparation}). The surrogate is a parameterised function $f_\theta$ that maps the inputs directly to the full saturation sequence,
\begin{equation}
\hat{y} = f_\theta(x), \qquad \hat{y} \approx y,
\label{eq:f_theta}
\end{equation}
where $\theta$ denotes the network parameters, learned by minimising the loss described in Section \ref{sec:surrogate_protocol}. All reporting times are predicted jointly, and the prediction at one reporting time therefore does not depend on the prediction at an earlier time.

\subsection{Data Preparation} \label{sec:data_preparation}

All fields are downsampled from the $842 \times 120$ simulation grid to $256 \times 64$ cells using bilinear interpolation, which coarsens the resolution from 10\,m $\times$ 10\,m to approximately 33\,m $\times$ 19\,m. This reduction lowers the computational cost of training while preserving the large-scale structure of the plume. At this resolution, thin layers and fault cores span only one or two cells, and their geometry is therefore represented only approximately. Both the training targets and the evaluation metrics are defined at this resolution, so the reported errors do not account for the error introduced by downsampling. The injection well is encoded as a separate input field, the well mask. In this field, a $3 \times 3$-cell patch centred on the well location takes the value of the injection rate, 0.035\,kg\,s$^{-1}$, and all other cells are zero. Since the injection rate is the same in all cases, the field effectively encodes the well location.

Porosity, permeability, and the initial pressure and temperature fields are normalised to $[0,1]$ using min-max normalisation,
\begin{equation}
\tilde{v} = \frac{v - v_{\min}}{v_{\max} - v_{\min}},
\label{eq:minmax_norm}
\end{equation}
where $v_{\min}$ and $v_{\max}$ denote the minimum and maximum values of the field in the training split. Since permeability spans several orders of magnitude, it is first transformed to logarithmic space, and Equation \ref{eq:minmax_norm} is applied to $\log_{10} K$. The resulting training statistics are then used to normalise the validation, test, and structural-shift data. The well mask is left non-normalised, and gas saturation is clipped to its physical range of $[0,1]$.

The data are split by structural geometry, with 195, 24, and 25 geometries assigned to the training, validation, and test splits, respectively, corresponding to 4,018, 495, and 487 simulation cases. Each split is augmented by reflecting every case about the vertical axis, a transformation that preserves the orientation of the layering relative to gravity and requires no additional simulation. Each case and its reflection remain in the same split, and both are included when evaluating the test split. The structural-shift subsets are evaluated without reflection.

\subsection{Architecture, Loss, and Training Protocol} \label{sec:surrogate_protocol}

The network follows a U-Net architecture, with skip connections between corresponding encoder and decoder levels. The encoder has four levels with 32, 64, 128, and 256 channels, respectively. Each level applies two $3 \times 3$ convolutions, each followed by group normalisation \citep{wu-2018} and Rectified Linear Unit (ReLU) activation, before $2 \times 2$ max pooling. The bottleneck uses two further convolutions with 512 channels, followed by two residual blocks. Each residual block comprises two convolutions with group normalisation and a residual connection around the convolutional pair. The decoder mirrors the encoder, using transposed convolutions for upsampling and concatenating the corresponding encoder features at each level. A final $1 \times 1$ convolution followed by a sigmoid activation constrains the predicted saturation to $[0,1]$.

The training loss combines a mean squared error term over all reporting times with an additional term that emphasises errors near the plume boundary. The mean squared error is weighted according to the stage of plume evolution,
\begin{equation}
\mathcal{L}_{\text{MSE}} = \frac{1}{T} \sum_{t=1}^{T} w_t \, \mathbb{E}\big[ (\hat{y}_t - y_t)^2 \big],
\label{eq:weighted_mse}
\end{equation}
where $T = 17$ and $w_t$ decreases linearly from 2.0 to 1.5 over the injection period, from 1.5 to 1.0 over the early post-injection period up to 200 years, and from 1.0 to 0.5 at later reporting times. This weighting assigns greater importance to errors during injection, when the plume evolves more rapidly, and reduces the contribution of later reporting times, when plume evolution is slower.

The second component weights the squared error by the local gradient magnitude of the target saturation,
\begin{equation}
\mathcal{L}_{\text{edge}} = \mathbb{E}\left[
\frac{\lVert\nabla y\rVert}{\overline{\lVert\nabla y\rVert} + \varepsilon}
\odot (\hat{y} - y)^2
\right],
\label{eq:edge_loss}
\end{equation}
where $\lVert\nabla y\rVert$ is the spatial gradient magnitude of the target saturation, computed from forward differences between neighbouring cells, and $\overline{\lVert\nabla y\rVert}$ is its mean over all cases, reporting times, and cells in the batch. The constant $\varepsilon = 10^{-6}$ prevents division by zero. Since every target contains CO$_2$ at all reporting times, the batch mean remains well above $\varepsilon$, and the weights have a mean of approximately one. This term increases the contribution of errors near the plume boundary, where saturation gradients are largest, relative to errors in regions where saturation varies more gradually. The total loss is
\begin{equation}
\mathcal{L} = \mathcal{L}_{\text{MSE}} + \lambda \, \mathcal{L}_{\text{edge}}, \qquad \lambda = 0.1.
\label{eq:total_loss}
\end{equation}

The network is trained using the AdamW optimiser \citep{kingma-2014, loshchilov-2019}, with a learning rate of $10^{-4}$, a weight decay of $10^{-2}$, a batch size of 64, and gradient clipping at a maximum norm of 1.0. Training runs for up to 100 epochs, with early stopping after 15 consecutive epochs without improvement in validation loss. The model with the lowest validation loss is retained for evaluation.

\begin{figure}[htbp]
    \centering
    \includegraphics[width=0.75\textwidth]{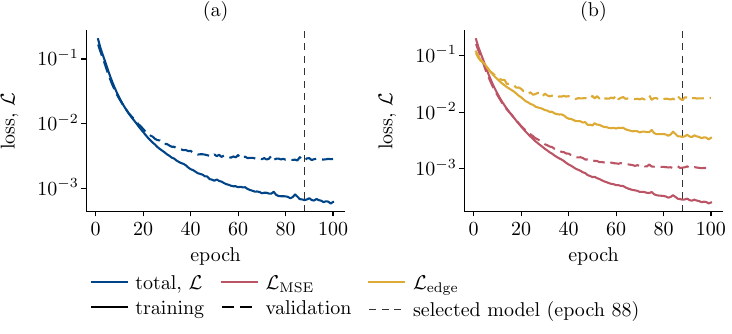}
    \caption{Training (solid) and validation (dashed) losses over 100 epochs: (a) total loss (Equation \ref{eq:total_loss}) and (b) the component losses, $\mathcal{L}_{\text{MSE}}$ and $\mathcal{L}_{\text{edge}}$. The vertical axis is logarithmic.}
    \label{fig:training_curves}
\end{figure}

Figure \ref{fig:training_curves} shows the training and validation losses over the course of training. Both decrease rapidly during the first 20 epochs, after which the validation loss remains above the training loss. This gap is most pronounced in the edge loss: the validation edge loss shows minimal change after approximately 30 epochs, while the training edge loss continues to decrease. The lowest validation loss occurs at epoch 88, and training therefore runs for the full 100 epochs, as fewer than 15 epochs elapse after this minimum.

A typical OPM Flow simulation takes approximately 40 minutes to complete, with runtimes ranging from 29 to 93 minutes, using 8 MPI processes with 2 threads each on a workstation with 64 logical cores. In contrast, the trained surrogate model, which has 17.2 million parameters, predicts the downsampled gas saturation sequence for a single case in approximately 0.015 s on an 18-core CPU, excluding data loading and preprocessing. Since the surrogate only predicts gas saturation at a reduced spatial resolution, this comparison is indicative rather than strictly equivalent. Training the surrogate model takes approximately 45 minutes on a single NVIDIA TITAN RTX GPU, also excluding data loading and preprocessing.

\subsection{In-Distribution Evaluation} \label{sec:surrogate_interpolation_results}

\begin{figure}[htbp]
    \centering
    \includegraphics[width=\textwidth]{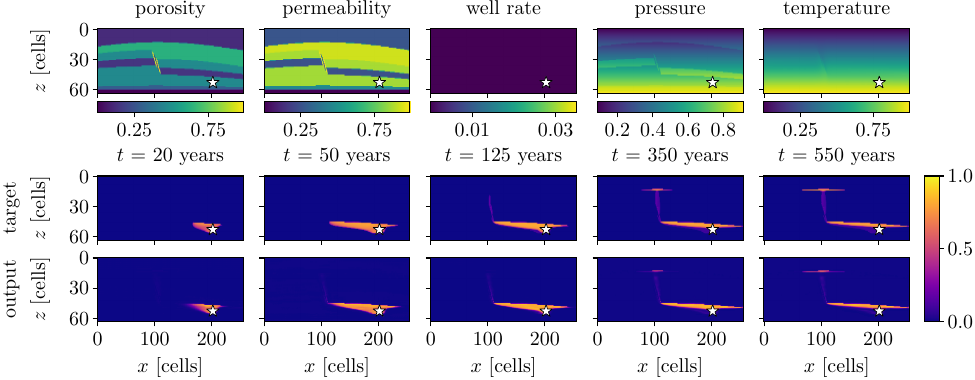}
    \caption{Inputs and results for the test case with the median plume-masked MAE. Top row: the five input fields, namely porosity, permeability, the well mask (\textit{well rate}, in kg\,s$^{-1}$), and the pressure and temperature fields at $t = 0$, with all fields except the well mask in normalised units. Middle and bottom rows: target and predicted gas saturation at five reporting times, from 20 to 550 years. Axes show cell indices of the downsampled $256 \times 64$ grid, and stars mark the injection well.}
    \label{fig:case_overview}
\end{figure}

The full input-to-output pipeline for the test case with the median plume-masked MAE, defined below, is shown in Figure \ref{fig:case_overview}. The porosity and permeability fields show the layering and a fault, and the initial pressure and temperature fields increase smoothly with depth, consistent with the hydrostatic and geothermal gradients. In the target, CO$_2$ spreads laterally from the well beneath a low-permeability layer during injection and, after injection, rises along the fault into a shallower layer. The surrogate reproduces the lateral spreading and the migration along the fault, but predicts less CO$_2$ accumulating in the shallower layer at late times.

Three plume-masked metrics are computed at each reporting time: the mean absolute error, the signed bias, and the relative $L_2$ error. These are restricted to cells where CO$_2$ is present in either the prediction or the target, preventing the large unaffected part of the domain from dominating the averages, following \citet{wen-2023} and \citet{ju-2024}. With $\hat{y}_{t,i,j}$ and $y_{t,i,j}$ denoting the predicted and target gas saturation in cell $(i, j)$ at reporting time $t$, the plume mask is
\[
M_t = \{(i,j) : \hat{y}_{t,i,j} > \tau \text{ or } y_{t,i,j} > \tau\}, \qquad \tau = 0.01.
\]
Within this mask, the mean absolute error and signed bias are
 \begin{equation}
    \text{MAE}_t = \frac{1}{|M_t|} \sum_{(i,j) \in M_t} \big| \hat{y}_{t,i,j} - y_{t,i,j} \big|,
    \qquad
    \text{bias}_t = \frac{1}{|M_t|} \sum_{(i,j) \in M_t} \big( \hat{y}_{t,i,j} - y_{t,i,j} \big),
    \label{eq:plume_metrics}
\end{equation}
where $|M_t|$ is the number of cells in the mask. The bias retains the sign of the error, and positive and negative errors can therefore cancel. The relative $L_2$ error is
\begin{equation}
    e_t = \frac{\|\hat{y}_t - y_t\|_{M_t}}{\big(\|y_t\|_{M_t}^2 + \varepsilon\big)^{1/2}},
    \qquad
    \|v\|_{M_t} = \Big( \sum_{(i,j) \in M_t} v_{i,j}^2 \Big)^{1/2},
    \label{eq:plume_rel_l2}
\end{equation}
where $\varepsilon = 10^{-6}$ ensures numerical stability.

\begin{figure}[htbp]
    \centering
    \includegraphics[width=\textwidth]{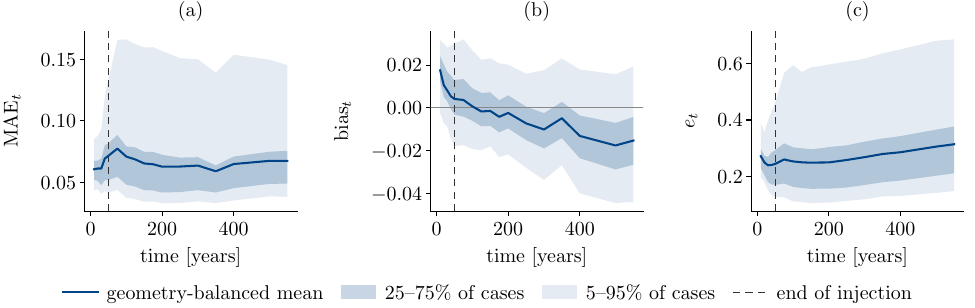}
    \caption{Plume-masked (a) MAE, (b) bias, and (c) relative $L_2$ error for the test set: geometry-balanced mean (line) and 25--75\% and 5--95\% ranges across the 974 test cases (bands). The dashed line marks the end of injection.}
    \label{fig:test_metrics}
\end{figure}

Test geometries contribute different numbers of cases, and each metric is therefore averaged first over the cases of each test geometry $g$, including their reflections, and then across the $N_g = 25$ test geometries,
\begin{equation}
\overline{\text{MAE}}_t = \frac{1}{N_g} \sum_{g=1}^{N_g} \big\langle \text{MAE}_t \big\rangle_g,
\label{eq:geo_balanced}
\end{equation}
and equivalently for $\overline{\text{bias}}_t$ and $\overline{e}_t$. Figure \ref{fig:test_metrics} shows these geometry-balanced metrics at all reporting times.

Averaged over all reporting times, $\overline{\text{MAE}}_t$ is 0.066, $\overline{\text{bias}}_t$ is $-0.002$, and $\overline{e}_t$ is 0.26. $\overline{\text{MAE}}_t$ increases from 0.061 during early injection to a maximum of 0.077 at 75 years, shortly after injection ends, and then decreases, remaining between 0.059 and 0.067 from 150 years onwards. $\overline{\text{bias}}_t$ follows a systematic trend. The surrogate over-predicts gas saturation within the plume during injection, with a bias of 0.018 at 10 years, and under-predicts it at late times, with the bias changing sign at about 125 years and reaching $-0.017$ at 500 years. Outside the plume mask, the predicted saturation is small but non-zero, as the sigmoid activation in the output layer is strictly positive.

\begin{figure}[htbp]
    \centering
    \includegraphics[width=\textwidth]{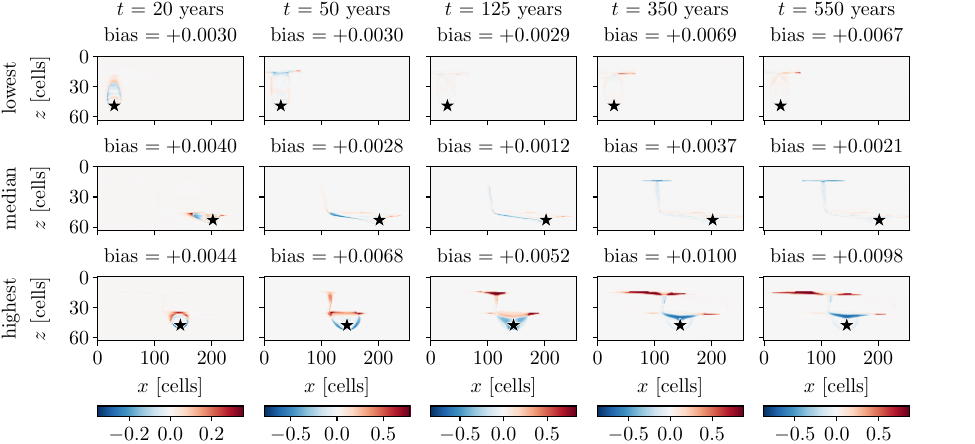}
    \caption{Signed gas saturation error (prediction minus target) for the test cases with the lowest, median, and highest plume-masked MAE (rows) at five reporting times (columns). Values above each panel are domain-averaged signed errors, which, unlike the plume-masked bias, include the small non-zero predictions outside the plume. Axes as in Figure \ref{fig:case_overview}, and stars mark the injection wells.}
    \label{fig:spatial_error}
\end{figure}

To examine errors spatially, three test cases are selected by their plume-masked MAE averaged over all reporting times: those with the lowest, median, and highest error. Figure \ref{fig:spatial_error} shows the signed error for these cases at five reporting times. In the lowest- and median-error cases, errors are concentrated in narrow bands along the plume boundary and along the fault through which CO$_2$ rises, with little error in the plume interior or the surrounding domain. In the highest-error case, a band of over-prediction beneath a shallower low-permeability layer is already visible at the end of injection and grows from 125 years onwards. A band of under-prediction develops along a deeper layer at the same time. The surrogate therefore moves too much CO$_2$ upwards in this case, rather than misplacing only the plume boundary. Across all three cases, errors remain largest near the plume boundary, where saturation gradients are steepest, despite the increased weighting of these regions in $\mathcal{L}_{\text{edge}}$ during training.

\begin{figure}[htbp]
    \centering
    \includegraphics[width=0.7\textwidth]{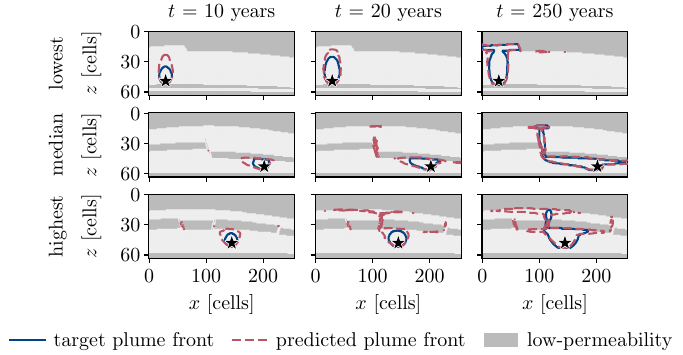}
    \caption{Target (solid blue) and predicted (dashed red) plume fronts, defined by $S_g = 0.01$, for the test cases with the lowest, median, and highest plume-masked MAE (rows) at 10, 20, and 250 years (columns). Low-permeability regions are shown in grey, and axes are as in Figure \ref{fig:case_overview}.}
    \label{fig:plume_front}
\end{figure}

As a complementary check on plume shape, Figure \ref{fig:plume_front} compares the target and predicted plume fronts, defined by the threshold $\tau$, for the same three cases at 10, 20, and 250 years. For the lowest- and median-error cases, the predicted fronts closely follow the target at 10 and 20 years. At 250 years, the main difference is the lateral extent of CO$_2$ beneath the caprock in the lowest-error case. For the highest-error case, the fronts agree at 10 years, but by 20 years the predicted front already extends above the target plume, beneath the shallower low-permeability layer. At 250 years, the predicted front reaches shallower layers that the target plume does not reach. The larger error of this case therefore reflects a difference in plume footprint rather than in boundary position alone. A comparison of predicted and target plume extent would complement the plume-masked metrics but is not computed here.

\subsection{Evaluation Under Structural Shift} \label{sec:surrogate_extrapolation_results}

Two structural-shift subsets assess surrogate performance on geological structures beyond those represented in the training data. Each subset comprises 50 simulation cases across 18 structural geometries. Both subsets are described in detail in Appendix \ref{app:shift_subsets}.

\begin{figure}[htbp]
    \centering
    \includegraphics[width=\textwidth]{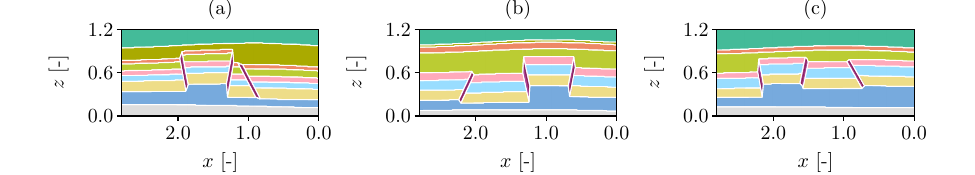}
    \caption{Example structural geometries from the extended-range subset, each with a fault pair and a supplementary fault. Colours and coordinates as in Figure \ref{fig:fault_styles}.}
    \label{fig:ood_subset_1_examples}
\end{figure}

The extended-range subset uses the same structural generator, with some parameter ranges extended beyond those used for training and others restricted to part of their training range (Table \ref{tab:ood_1_stats}). The faulted sequence includes more and thinner layers with near-flat horizons, and the basal unit is divided into only two layers. Every geometry has a fault pair with a raised central block and a supplementary fault, a configuration present in only 7\% of the geometries in the main dataset. The fault pairs are also wider than in training, giving a broader central block, and their two faults have independent rather than shared throws. The fault tips lie within a narrower and lower range of elevations. Figure \ref{fig:ood_subset_1_examples} shows three example geometries.

\begin{figure}[htbp]
    \centering
    \includegraphics[width=\textwidth]{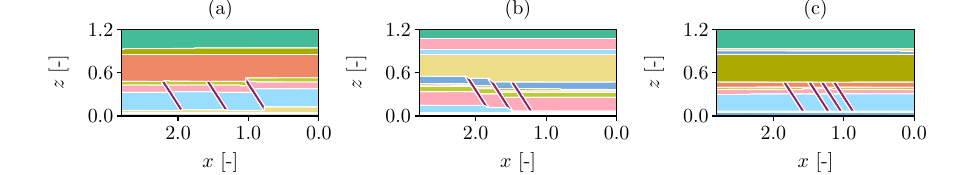}
    \caption{Example structural geometries from the parallel-fault subset, each with two to four parallel faults within a faulted region bounded by unfaulted layers. Colours and coordinates as in Figure \ref{fig:fault_styles}.}
    \label{fig:ood_subset_2_examples}
\end{figure}

Figure \ref{fig:ood_subset_2_examples} shows three example geometries from the parallel-fault subset.

This subset is built with a separate generator and introduces a fault topology that the training configuration of the structural generator cannot produce. Two to four faults with a common dip cut a faulted region bounded above and below by unfaulted layers, and each fault has an independently sampled throw and downthrown side. The faults therefore divide the faulted region into a sequence of blocks with independent offsets, rather than the single central block of a fault pair. The faulted region lies either low in the domain, beneath a thick unfaulted sequence, or high, beneath a thin one. The common dip of 25--36$^\circ$ is shallower than that of the main faults in the training data but similar to that of the supplementary faults.

\begin{figure}[htbp]
    \centering
    \includegraphics[width=\textwidth]{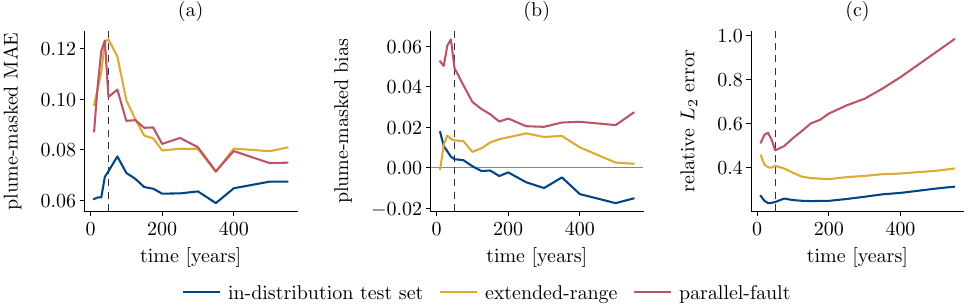}
    \caption{Geometry-balanced plume-masked (a) MAE, (b) bias, and (c) relative $L_2$ error for the test set and the two structural-shift subsets. The dashed line marks the end of injection.}
    \label{fig:shift_metrics}
\end{figure}

The surrogate is evaluated on both subsets using the plume-masked metrics of Equations \ref{eq:plume_metrics} and \ref{eq:plume_rel_l2}, averaged first within each structural geometry and then across the 18 geometries of each subset (Equation \ref{eq:geo_balanced}). Figure \ref{fig:shift_metrics} compares these metrics with those of the test set.

Both structural shifts increase prediction error relative to the test set. Averaged over all reporting times, $\overline{\text{MAE}}_t$ increases from 0.066 for the test set to 0.094 for the extended-range subset and 0.091 for the parallel-fault subset, and $\overline{e}_t$ increases from 0.26 to 0.38 and 0.65, respectively. As for the test set, $\overline{\text{MAE}}_t$ is largest around the end of injection. For the extended-range subset, it rises from 0.098 at 10 years to 0.124 at 50 years, compared with 0.071 for the test set, and then decreases to about 0.08 from 200 years onwards. For the parallel-fault subset, it peaks earlier, at 0.123 at 40 years, and decreases to 0.075 at the final reporting time, compared with 0.067 for the test set.

The relative $L_2$ error separates the two subsets more clearly. For the extended-range subset, $\overline{e}_t$ decreases from 0.46 at 10 years to between 0.35 and 0.41 for the remainder of the simulation, between 0.08 and 0.18 above the test set. For the parallel-fault subset, $\overline{e}_t$ remains between 0.48 and 0.56 during injection and then increases steadily to 0.98 at the final reporting time, while $\overline{\text{MAE}}_t$ decreases. The plume mask includes cells where either the prediction or the target exceeds the saturation threshold. A growing relative error alongside a decreasing absolute error is therefore consistent with an increasing mismatch between the predicted and target plume footprints.

\begin{figure}[htbp]
    \centering
    \includegraphics[width=\textwidth]{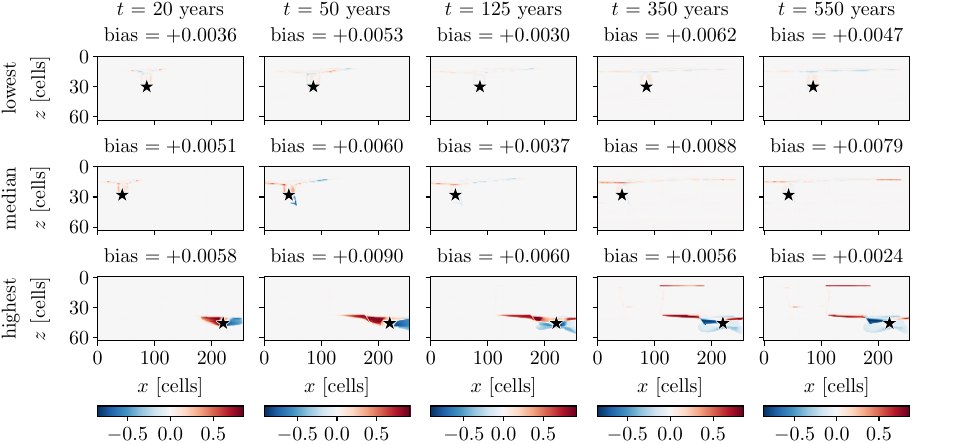}
    \caption{Signed gas saturation error for the extended-range cases with the lowest, median, and highest plume-masked MAE, as in Figure \ref{fig:spatial_error}.}
    \label{fig:error_extended}
\end{figure}

Both subsets show a positive bias, in contrast to the test set, in which the bias changes from positive to negative after injection. For the extended-range subset, $\overline{\text{bias}}_t$ is positive at all but the first reporting time and reaches at most 0.017. For the parallel-fault subset, it is positive at every reporting time, between 0.020 and 0.063, and largest during injection. The surrogate therefore systematically over-predicts gas saturation for the parallel-fault geometries. To examine these errors spatially, the cases with the lowest, median, and highest plume-masked MAE averaged over reporting times are selected for each subset, as for the test set. Figures \ref{fig:error_extended} and \ref{fig:error_parallel} show the signed error for these cases, and their input fields are shown in Appendix \ref{app:shift_subsets}.

For the extended-range subset, errors in the lowest- and median-error cases are small, as in the test set. These are confined to the plume boundary near the well and to thin horizontal bands where CO$_2$ spreads beneath low-permeability layers. The highest-error case differs qualitatively. From the first reporting time onwards, a band of over-prediction extends from the well along the layer containing it, next to a region of under-prediction around the well. This pattern persists after injection, indicating that the predicted plume is displaced relative to the target rather than differing only at its boundary.

\begin{figure}[htbp]
    \centering
    \includegraphics[width=\textwidth]{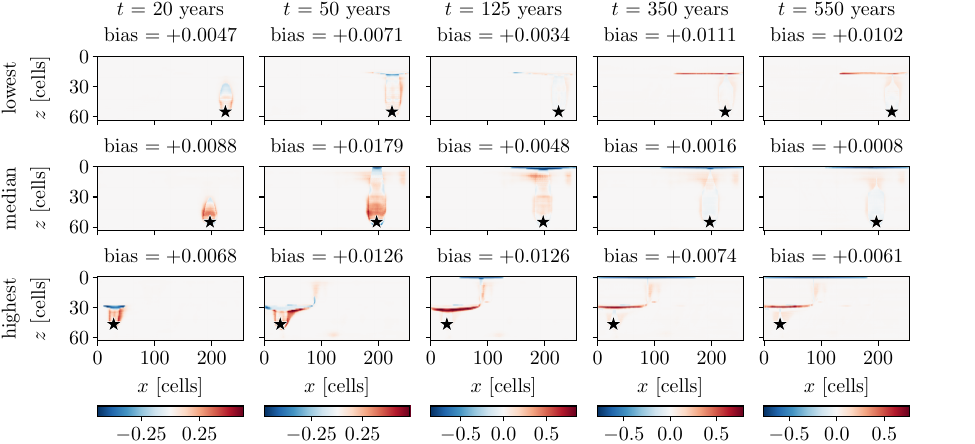}
    \caption{Signed gas saturation error for the parallel-fault cases with the lowest, median, and highest plume-masked MAE, as in Figure \ref{fig:spatial_error}.}
    \label{fig:error_parallel}
\end{figure}

For the parallel-fault subset, errors in the lowest-error case occur at the boundary of the rising plume during injection and, after injection, along the first low-permeability layer that the plume reaches. In the median case, the surrogate over-predicts saturation in the rising plume during injection and under-predicts the accumulation beneath the caprock shortly afterwards. In the highest-error case, the well lies beside the faulted region, beneath a thin low-permeability layer. From the end of injection onwards, a band of over-prediction develops beneath this layer and extends laterally towards the faulted region, indicating that the surrogate spreads CO$_2$ further beneath the layer than the simulation does. This lateral over-prediction is consistent with the positive bias and growing relative error of this subset.

Together, these results show that both structural shifts reduce surrogate accuracy, with the largest effect on the relative $L_2$ error for the parallel-fault subset. Each subset changes several factors jointly. The extended-range subset changes the number and thickness of layers, the fault configuration, the fault-pair width and throws, the fault tip elevation, and the proportion of sealing fault cores. The parallel-fault subset combines a new fault topology with more faults and layers than the training geometries and contains no sealing fault cores. Well depths in both subsets also differ from those in the training data (Appendix \ref{app:shift_subsets}). These results therefore do not attribute the loss of accuracy to any individual factor.

\section{Discussion} \label{sec:discussion} 

\textit{FlowersBench} is designed to evaluate how well surrogates generalise across geological structures. Existing benchmark simulation datasets for GCS assess generalisation within a fixed structural geometry (Section \ref{sec:introduction}), whereas \textit{FlowersBench} provides a population of distinct structural geometries. We evaluate the baseline in two settings: on unseen geometries sampled using the training configuration of the structural generator, and on two subsets with systematic differences from the training geometries.

In the lowest- and median-error test cases, the surrogate reproduces the main migration pathways, including lateral spreading beneath low-permeability layers and upward flow along fault damage zones. These pathways depend on stratigraphy and fault configuration, which the surrogate receives through the porosity and permeability fields, while the initial pressure and temperature fields encode the hydrostatic and geothermal gradients at the start of injection. In all examined test cases, the largest errors occur near the plume boundary, consistent with the patterns reported by \citet{badawi-2025}. Across the test set, the plume-masked bias changes from over-prediction during injection to under-prediction at late times (Section \ref{sec:surrogate_interpolation_results}).

The structural-shift evaluation examines whether surrogate performance is maintained when the geological structure differs from that represented in the training data. Both structural shifts reduce accuracy, and the error evolves differently in the two subsets. For the extended-range subset, the plume-masked MAE increases from 0.066 on the test set to 0.094, while the relative $L_2$ error increases from 0.26 to 0.38. The temporal pattern of the MAE and relative $L_2$ error remains similar to that of the test set, indicating that extending the parameter ranges primarily increases the magnitude of the error. The parallel-fault subset has a similar plume-masked MAE of 0.091, whereas the relative $L_2$ error increases steadily after injection, reaching 0.98 at the final reporting time. The surrogate also systematically over-predicts gas saturation in this subset. Together, these results are consistent with an increasing mismatch between the predicted and simulated plume footprints during the post-injection period, as CO$_2$ migrates through the faulted region and beneath low-permeability layers.

Each subset varies several factors jointly (Section~\ref{sec:surrogate_extrapolation_results}), and their individual effects therefore cannot be isolated. Defining subsets in which these factors vary independently would allow their individual contributions to be assessed. The structural generators used here would allow such single-factor subsets to be constructed in future work.

These results characterise surrogate behaviour under specific structural shifts rather than general performance beyond the training distribution. A benchmark supports conclusions only about the conditions it represents \citep{shirali-2023}, and GCS surrogate evaluation therefore requires geological structures that reflect the range of conditions relevant to the intended application. The FluidFlower study provides a precedent through its combination of a realistic setting, explicit validation objectives, and a transparent experimental process \citep{fluidflower-2024}. Following this precedent, \textit{FlowersBench} extends a FluidFlower-inspired setting from a single reference geometry to a family of structural geometries, with explicitly defined evaluation settings and openly released data. Surrogate generalisation can therefore be evaluated across systematic changes in geological structure.

\section{Limitations and Future Directions} \label{sec:limitations}

The range of geological structures in \textit{FlowersBench} is bounded by the structural generator. The generator is two-dimensional and procedural, and does not represent erosional unconformities, branching fault networks, curved faults, or fully three-dimensional structures \citep{wellmann-2018}. The parameter ranges are chosen to span plausible structural variation within the settings of FluidFlower and SPE11B \citep{fluidflower-2024, nordbotten-2024} and are not calibrated against field data or expert interpretations. The resulting variability therefore differs from the interpretive uncertainty documented by \citet{bond-2015}, in which experts derive different structural models from the same observations. Moreover, the 244 structural geometries form a finite sample from the generator, and the effect of the number of training geometries on generalisation has not been assessed.

The simulation settings are fixed across all cases, with a single injection well, a fixed injection rate and schedule, and OPM Flow as the sole simulator. Modelling and simulator choices can lead to notable differences in predicted storage behaviour \citep{fluidflower-2024}, and this source of variability is not represented in the dataset. The simulations also exclude geomechanical coupling, and the pressure increases reported in Section \ref{sec:dataset_characterisation} are therefore not assessed against caprock or fault stability. 

The baseline surrogate has three further limitations, related to resolution, training, and the prediction task. The input and output fields are downsampled from $842 \times 120$ to $256 \times 64$ cells, at which thin layers and fault cores span only one or two cells (Section \ref{sec:data_preparation}). The contribution of the loss weights and $\lambda$ to accuracy is not assessed through ablation. The evaluation considers a single architecture and gas saturation as the sole prediction target, without estimates of predictive uncertainty.

These limitations motivate several directions for future work. The structural generator could be extended, for example to three-dimensional structures, and its parameter ranges calibrated against expert-derived or inversion-derived structural ensembles \citep{wellmann-2018}. Varying injection rates, well configurations, and injection schedules would extend the evaluation beyond geological structure. Training on increasing numbers of structural geometries would show whether dataset size limits generalisation. Alternative architectures, including graph-based models that represent structural geometry directly \citep{ju-2024}, can be used to test whether explicit representations of geological structure improve generalisation. Finally, extending the prediction task to pressure and dissolved CO$_2$ would broaden the range of dynamic responses evaluated, and estimates of predictive uncertainty would complement the error metrics reported here.

\section{Conclusion}

This work introduced \textit{FlowersBench} to evaluate how GCS surrogates generalise across geological structures, which existing benchmark simulation datasets hold fixed. The 5,000 simulation cases span 244 structural geometries from an in-house structural generator, with stratigraphic layering and fault configurations varying between geometries, and property realisations and well locations varying within each geometry. The dataset supports two evaluation settings. In-distribution evaluation uses test geometries withheld from training. Evaluation under structural shift uses two subsets. The extended-range subset is generated with a modified configuration of the structural generator, and the parallel-fault subset with a separate generator that produces an unseen fault topology.

A baseline U-Net surrogate, trained to predict gas saturation at all reporting times, illustrates how the two settings differ. For the extended-range subset, the error follows a similar temporal pattern to that of the test set, indicating that extending the parameter ranges primarily increases the magnitude of the error. For the parallel-fault subset, the relative $L_2$ error increases steadily after injection, and the surrogate systematically over-predicts gas saturation. Each subset varies several factors jointly, and these results therefore characterise surrogate behaviour under the specific structural shifts considered rather than general performance beyond the training distribution.

Alternative architectures and prediction targets can be assessed on \textit{FlowersBench} under consistent conditions, both on unseen geometries and under structural shift. By treating geological structure as an explicit dimension of evaluation, \textit{FlowersBench} complements existing benchmark simulation datasets and supports the development of surrogates for storage settings where geological structure is uncertain.

\section*{Data Availability}

The \textit{FlowersBench} dataset is available at \url{https://doi.org/10.5281/zenodo.24380327} under the \href{https://creativecommons.org/licenses/by-nc/4.0/}{CC BY-NC 4.0} licence. It contains the geometry files of the 244 structural geometries, the simulation and initial-state files of the 5,000 simulation cases, the property files of their 1,146 property realisations, and the corresponding files of the two structural-shift subsets (approximately 56\,GB in total). The file structure is described in Appendix \ref{app:data_records}, and the parameters of the structural generator are listed in Appendices \ref{app:generator-params} and \ref{app:shift_subsets}.

\section*{Acknowledgements}

This work is funded by the Engineering and Physical Sciences Research Council's ECO-AI Project grant (reference number EP/Y006143/1), with additional financial support from the PETRONAS Centre of Excellence in Subsurface Engineering and Energy Transition (PACESET).

\clearpage

\appendix
\section*{Appendix}
\addcontentsline{toc}{section}{Appendix}

\section{Governing Equations} \label{app:governing_equations}

The simulations follow the two-phase, two-component formulation with thermal effects defined for SPE11B \citep{nordbotten-2024}, summarised here. For the CO$_2$-rich, non-wetting gas phase, $\alpha = n$, and the H$_2$O-rich, wetting liquid phase, $\alpha = w$, the multiphase Darcy law is
\begin{equation}
\mathbf{u}_\alpha = -\frac{k_{r,\alpha} \mathbf{k}}{\mu_\alpha}\left(\nabla p_\alpha - \rho_\alpha \mathbf{g}\right),
\end{equation}
where $\mathbf{u}_\alpha$ is the volumetric flux of phase $\alpha$, $p_\alpha$ is the phase pressure, $k_{r,\alpha}$ and $\mathbf{k}$ are the relative and intrinsic permeabilities, respectively, $\mu_\alpha$ is the phase viscosity, $\rho_\alpha$ is the phase density, and $\mathbf{g}$ is the gravitational acceleration. Mass conservation for the components $i = \mathrm{CO_2}$ and $i = \mathrm{H_2O}$ is
\begin{equation}
\sum_{\alpha=w,n}\left[\frac{\partial}{\partial t}\Big(\rho_\alpha \phi \, s_\alpha \, \chi^i_\alpha \, J\Big)
+ \nabla \cdot \Big(\rho_\alpha \mathbf{u}_\alpha \chi^i_\alpha + \rho_\alpha \mathbf{j}^i_\alpha\Big)\right] = 0,
\end{equation}
where $\phi$ is porosity, $s_\alpha$ is the saturation of phase $\alpha$, $\chi^i_\alpha$ is the mass fraction of component $i$ in phase $\alpha$, and $J$ is the reservoir volume per domain volume, which adds volume at the lateral boundaries. The diffusive and dispersive flux is
\begin{equation}
\mathbf{j}^i_\alpha = -\left(s_\alpha \phi \, D_\alpha + E\left|\mathbf{u}_\alpha\right|\right) \nabla \chi^i_\alpha,
\end{equation}
where $D_\alpha$ is the diffusion coefficient in phase $\alpha$ and $E$ is the dispersivity. The phase pressures are related by the capillary pressure, $p_n - p_w = p_{\mathrm{cap}}(s_w)$, and the system is closed by
\begin{equation}
\sum_{\alpha=w,n} s_\alpha = 1, \qquad \sum_{i=\mathrm{CO_2},\mathrm{H_2O}} \chi^i_\alpha = 1.
\end{equation}
CO$_2$ dissolved in the liquid phase and water vapour in the gas phase are represented by non-zero mass fractions $\chi^{\mathrm{CO_2}}_w$ and $\chi^{\mathrm{H_2O}}_n$. The rock is incompressible, and porosity does not change over time.

Energy conservation, summed over the liquid, gas, and solid phases, is
\begin{equation}
\sum_{\alpha=w,n,s} \left[ \frac{\partial}{\partial t}\Big(\rho_\alpha \phi \, s_\alpha \, (e_\alpha - gz) \, J\Big)
+ \nabla \cdot \Big(\mathbf{u}_\alpha\big(\rho_\alpha e_\alpha + p - \rho_\alpha g z\big) - \phi \, s_\alpha \, \kappa_\alpha \nabla T\Big) \right] = 0,
\end{equation}
where $e_\alpha$ is the specific internal energy and $\kappa_\alpha$ the thermal conductivity of phase $\alpha$. For the solid phase, $\phi s_s = 1 - \phi$ and $\mathbf{u}_s = 0$, so that the solid contributes only heat storage and conduction, with internal energy $e_s = c_s T$ (Table~\ref{tab:fixed_properties}). CO$_2$ is injected at 10\,$^\circ$C, well below the reservoir temperature, and the resulting temperature changes affect fluid density, viscosity, and CO$_2$ solubility, coupling the thermal and flow processes.

Relative permeability and capillary pressure are defined in terms of the normalised saturations
\begin{equation}
s_{w,n} = \max\!\left(\frac{s_w - s_{w,\mathrm{imm}}}{1 - s_{w,\mathrm{imm}}}, 0\right), \qquad
s_{n,n} = \max\!\left(\frac{s_n - s_{n,\mathrm{imm}}}{1 - s_{n,\mathrm{imm}}}, 0\right),
\end{equation}
where $s_{\alpha,\mathrm{imm}}$ is the saturation below which phase $\alpha$ is immobile. Relative permeability follows the Brooks-Corey form $k_{r,\alpha} = s_{\alpha,n}^{\,c_1}$, and capillary pressure follows
\begin{equation}
\tilde{p}_{\mathrm{cap}}(s_w) = p_{\mathrm{entry}}\, s_{w,n}^{-1/c_2}, \qquad
p_{\mathrm{cap}}(s_w) = p_{\mathrm{cap,max}} \, \mathrm{erf}\!\left(\frac{\tilde{p}_{\mathrm{cap}}(s_w)}{p_{\mathrm{cap,max}}} \frac{\sqrt{\pi}}{2}\right),
\end{equation}
where the error function limits the capillary pressure smoothly to $p_{\mathrm{cap,max}}$. The exponents $c_1 = c_2 = 1.5$ are the same for all facies, as are $s_{n,\mathrm{imm}}$ and $p_{\mathrm{cap,max}}$ (Table~\ref{tab:fixed_properties}), whereas $s_{w,\mathrm{imm}}$ and $p_{\mathrm{entry}}$ are sampled for each region (Section~\ref{sec:property_sampling}).

The domain is initially saturated with water and follows the geothermal gradient of SPE11B, which corresponds to a depth of about 2,000\,m to the top of the domain,
\begin{equation}
T_{\mathrm{geo}}(z) = 70 - 0.025\,z,
\end{equation}
with $T_{\mathrm{geo}}$ in $^\circ$C and $z$ the elevation above the base of the domain in metres, and with hydrostatic pressure set by 300\,bar at a depth of 900\,m below the top (Section~\ref{sec:simulation_setup}). All boundaries are closed to fluid flow. The lateral boundaries are thermally insulating, $\mathbf{n} \cdot \nabla T = 0$, and the top and bottom boundaries are held at $T_{\mathrm{geo}}$. Additional volume at the lateral boundaries, represented by $J$ and implemented by increasing the pore volume of the lateral boundary cells, emulates connection to a larger surrounding aquifer and limits the pressure increase in the domain.

\clearpage
\section{Data Records} \label{app:data_records}

\textit{FlowersBench} is released as geometry files, simulation files, initial-state files, and property files, both for the 5,000 simulation cases of the main dataset and for the two structural-shift subsets (Section \ref{sec:surrogate_extrapolation_results}). The full release totals approximately 56\,GB. The main dataset and the two subsets are stored in separate folders, \texttt{1\_main\_dataset}, \texttt{2\_extended\_range\_subset}, and \texttt{3\_parallel\_fault\_subset}, each containing the subfolders \texttt{geometry\_files}, \texttt{simulation\_files}, \texttt{initial\_state\_files}, and \texttt{property\_files}. Each folder also contains a list of simulation cases (\texttt{simulation\_cases.csv}), which assigns every case of the main dataset to the training, validation, or test split used in Section \ref{sec:surrogate_baseline}, and an overview sheet of all structural geometries (\texttt{geometry\_overview.pdf}). A README and a quick-start notebook are provided at the top level.

Each structural geometry is provided as a \texttt{.geo} file and an accompanying \texttt{.json} file, shared by all simulation cases of that geometry. The \texttt{.geo} file stores the polygon boundaries of all regions in normalised generator coordinates (Section \ref{sec:geometry_generation}), grouped into numbered physical surfaces. The \texttt{.json} file maps each surface number to a region label: \texttt{Layer\_k} for stratigraphic layers, numbered from the bottom of the domain upward, with \texttt{Layer\_0} the impermeable basal layer, and \texttt{FAULT\_k} and \texttt{FAULT\_k\_MID} for the damage zone and core of fault $k$, numbered in order of insertion. In the parallel-fault subset, fault cores are labelled \texttt{FAULT\_k\_CORE}.

Each simulation case is stored as a compressed NumPy archive named
\begin{center}
\texttt{geo\{geo\}\_run\{run\}\_well\{well\}.npz},
\end{center}
where the three indices identify the structural geometry, property realisation, and well location.
Geometry indices in the main dataset are zero-padded to three digits. Each archive contains static fields, dynamic fields, and metadata, with an average size of approximately 10.6\,MB. A companion file with the suffix \texttt{\_init} stores the pressure and temperature fields at $t = 0$, the end of the settling period (Section \ref{sec:simulation_setup}). Reflected cases used for surrogate training (Section \ref{sec:surrogate_baseline}) are generated by reversing the horizontal axis of all fields and are not distributed as separate files.

A property file for each property realisation, named \texttt{geo\{geo\}\_run\{run\}\_properties.npz}, stores the immobile wetting saturation (\texttt{swimm}), the capillary entry pressure in Pa (\texttt{pentry}), and the dispersivity in m (\texttt{disperc}) of every cell, as arrays of shape $(n_z, n_x) = (120, 842)$. These properties apply to all well locations of the property realisation.

The two structural-shift subsets are released in the same format. Each contains 50 simulation cases from 18 structural geometries, with geometry indices between 9000 and 9023 for the extended-range subset and between 5026 and 5187 for the parallel-fault subset. Most subset geometries have a single property realisation with one to four well locations, and each subset occupies approximately 0.5\,GB.

\subsection{Static fields}

Each simulation archive stores four static fields: porosity (\texttt{poro}), horizontal permeability in mD (\texttt{permx}), vertical permeability in mD (\texttt{permz}, equal to $0.1\,$\texttt{permx}), and rock thermal conductivity in kJ\,m$^{-1}$\,day$^{-1}$\,K$^{-1}$ (\texttt{thconr}), the unit used by OPM Flow, which converts to W\,m$^{-1}$\,K$^{-1}$ on division by 86.4. The companion initial-state file stores the pressure in bar (\texttt{pressure0}) and the temperature in $^\circ$C (\texttt{temp0}) at $t = 0$. All static and initial-state fields are arrays of shape $(n_z, n_x) = (120, 842)$. The first axis corresponds to depth, increasing from the top of the domain ($z = 0$) to the base ($z = 1200$\,m), and the second to horizontal position, increasing from $x = 0$ to $x = 8400$\,m.

The \texttt{.geo} files instead use generator coordinates, in which $x_{\mathrm{geo}}$ increases from right to left and $z_{\mathrm{geo}}$ is the elevation above the base of the domain. Positions in the two systems are related by
\[
(x, z) = \bigl(8400 - 3000\,x_{\mathrm{geo}},\ 1200 - 1000\,z_{\mathrm{geo}}\bigr)\ \mathrm{m}.
\]

\subsection{Dynamic fields}

Each simulation archive stores four dynamic fields: gas saturation (\texttt{sgas}), pressure in bar (\texttt{pressure}), temperature in $^\circ$C (\texttt{temp}), and dissolved CO$_2$ as the solution gas-water ratio in sm$^3$\,sm$^{-3}$ (\texttt{rsw}). Each field is stored as an array of shape $(17, n_z, n_x)$, with the spatial axes ordered as for the static fields.

The 17 reporting times are given in years after the start of injection. These include 10, 20, 30, 40, and 50 years during injection, 75, 100, 125, 150, 175, and 200 years in the early post-injection period, and 250, 300, 350, 400, 500, and 550 years in the later monitoring period. The stored times, \texttt{key\_years}, are measured from the end of the equilibration period and read 10.1, 20.1, 30.1, 40.1, 50.1, 75, 100, \ldots, 500, and 549.7. The exact simulator report times, in years of 365.25 days, are given in \texttt{years}.

\subsection{Metadata}

Each simulation archive also stores metadata. The case identifiers \texttt{geo}, \texttt{run}, and \texttt{well} match the file name, and \texttt{nx} and \texttt{nz} give the grid dimensions of 842 and 120. The array \texttt{years} lists all 113 simulator report times, from $-1000$ at the start of equilibration to 549.7, with the $t = 0$ state at 0.1. The array \texttt{key\_years} lists the 17 reporting times of the dynamic fields.

The injection cell is recorded in \texttt{well\_x} and \texttt{well\_z} as one-based grid indices $(I, K)$ rather than as physical coordinates. All cells are 10\,m high, and the depth of the cell centre is therefore $z = 10(K - 1) + 5$\,m. The two outermost columns on each side are 1\,m and 9\,m wide, and the horizontal position of an interior cell centre is $x = 10(I - 2) + 5$\,m for $3 \leq I \leq 840$.

The initial-state files store the same case identifiers, together with the simulation time (\texttt{time\_days}) and report step (\texttt{report\_step}) of the $t = 0$ state.

\clearpage
\section{Generator Parameter Specification} \label{app:generator-params}

Table \ref{tab:generator-params} lists the sampling distributions of the in-house structural generator for the 244 structural geometries in \textit{FlowersBench}. Lengths and elevations are given in normalised generator coordinates (Section \ref{sec:geometry_generation}), and dip angles are given as physical angles in the simulation domain.

\begin{table}[H]
\centering
\caption{Sampling distributions of the in-house structural generator. Integer sets are sampled uniformly. Lengths and elevations are in normalised generator coordinates; dips are physical angles.}
\label{tab:generator-params}
\small
\setlength{\tabcolsep}{5pt}
\renewcommand{\arraystretch}{1.12}
\begin{tabularx}{\linewidth}{
>{\raggedright\arraybackslash}p{0.27\linewidth}
>{\raggedright\arraybackslash}p{0.27\linewidth}
>{\raggedright\arraybackslash}X
}
\toprule
Parameter & Distribution / value & Description \\
\midrule
\multicolumn{3}{l}{\textit{Stratigraphy}} \\
Domain & $[0, 2.8] \times [0, 1.2]$ & Normalised generator domain \\
Initial horizons & $\{3, 4, 5\}$ & Horizons of the faulted sequence \\
Basal-unit horizons & $\{2, 3\}$ & Horizons subdividing the basal unit beneath the faulted sequence \\
Horizon amplitude, $a_k$ & $\mathrm{Uniform}(0.01, 0.10)$ & Fraction of the height of the region being divided \\
Horizon frequency, $f_k$ & $\mathrm{Uniform}(0.2, 0.5)$ & Sinusoidal cycles across the domain width \\
Horizon phase, $\psi_k$ & $\mathrm{Uniform}(-0.01, 0.01)$ & Phase shift (rad) \\
Base height fraction, $b_k$ & $\mathrm{Uniform}(0.10,\, u_k)$, $u_k = 0.21$--$0.25$ & Base height within the region being divided; $u_k$ increases by 5\% per horizon, and the first horizon is raised by 0.2 \\
\midrule
\multicolumn{3}{l}{\textit{Fault pair or single fault}} \\
Fault style & Single (0.30), raised central block (0.45), lowered central block (0.25) & Paired styles use two faults dipping towards each other \\
Centre location, $x_{\mathrm{loc}}$ & $\mathrm{Uniform}(0.8, 2.0)$ & Centre of the fault pair; a single fault is placed at $x_{\mathrm{loc}} + w/2$ \\
Fault-pair width, $w$ & $\mathrm{Uniform}(0.15, 0.30)$ & Horizontal separation of the fault bases \\
Dip angle, $\gamma$ & $\mathrm{Uniform}(55^\circ, 65^\circ)$ & Sampled independently for each fault \\
Tip elevation, $z_{\mathrm{top}}$ & $\mathrm{Uniform}(0.79, 1.19)$ & Upper termination of the faults \\
Throw, $s$ & $\mathrm{Uniform}(0.02, 0.20)$ & Shared by both faults of a pair \\
Downthrown side & Left or right, equal probability & Single-fault style only \\
Damage-zone thickness & $0.050$ & Outer fault zone \\
Fault-core thickness & $0.015$ & Inner zone within the damage zone \\
\midrule
\multicolumn{3}{l}{\textit{Supplementary fault (paired styles only)}} \\
Probability & $0.30$ & Inserted if a valid location is found \\
Location & $\mathrm{Uniform}(0.2, 2.6)$ & At least 0.2 from the fault pair \\
Dip angle & $25^\circ$--$36^\circ$ & Dip direction sampled with equal probability \\
Tip elevation & $\mathrm{Uniform}(0.39, 0.89)$ & Upper termination of the fault \\
Throw & $\mathrm{Uniform}(0.02, 0.05)$ & Downthrown side sampled with equal probability \\
\bottomrule
\end{tabularx}
\end{table}

\clearpage
\section{Structural-Shift Subsets} \label{app:shift_subsets}

The two structural-shift subsets test how the surrogate responds to structural geometries outside the training configuration of the structural generator. Each subset includes 50 simulation cases from 18 structural geometries. The subsets also differ from the main dataset in other characteristics, including the number of permeable layers, the sealing behaviour of faults, and the depth at which wells are located. Table \ref{tab:subset_summary} summarises these characteristics for both subsets and for the main dataset.

\begin{table}[htbp]
\centering
\caption{Characteristics of the two structural-shift subsets compared with the main dataset.}
\label{tab:subset_summary}
\renewcommand{\arraystretch}{1.2}
\begin{tabular}{lccc}
\toprule
 & Main dataset & Extended-range & Parallel-fault \\
\midrule
Structural geometries & 244 & 18 & 18 \\
Property realisations & 1,146 & 20 & 18 \\
Simulation cases & 5,000 & 50 & 50 \\
Stratigraphic layers per geometry & 6--9 & 8--10 & 9--13 \\
Faults per geometry & 1--3 & 3 & 2--4 \\
Permeable layers per property realisation & 3--6 & 5--7 & 6--10 \\
Mean permeability of permeable layers (mD) & 271.6 & 411.1 & 552.6 \\
Sealing fault cores & 30.7\% & 45.0\% & 0\% \\
Well depth, median (IQR) (m) & 905 (835--945) & 635 (475--803) & 1,025 (808--1,045) \\
Plume extent at 550 years, mean (km$^2$) & 0.60 & 0.50 & 0.45 \\
\bottomrule
\end{tabular}
\end{table}

\subsection{Extended-range subset}

\begin{table}[htbp]
\centering
\caption{Generator parameters changed for the extended-range subset, relative to the main dataset (Table \ref{tab:generator-params}).}
\label{tab:ood_1_stats}
\small
\renewcommand{\arraystretch}{1.15}
\begin{tabularx}{\linewidth}{>{\raggedright\arraybackslash}X >{\raggedright\arraybackslash}p{0.27\linewidth} >{\raggedright\arraybackslash}p{0.27\linewidth}}
\toprule
Parameter & Main dataset & Extended-range subset \\
\midrule
Initial horizons & $\{3, 4, 5\}$ & $\{6, 7, 8\}$ \\
Horizon amplitude, $a_k$ & $\mathrm{Uniform}(0.01, 0.10)$ & $\mathrm{Uniform}(0.008, 0.010)$ \\
Horizon frequency, $f_k$ & $\mathrm{Uniform}(0.2, 0.5)$ & $\mathrm{Uniform}(0.5, 0.8)$ \\
Horizon phase, $\psi_k$ & $\mathrm{Uniform}(-0.01, 0.01)$ & $\mathrm{Uniform}(-0.02, -0.01)$ \\
Lower bound of base height fraction & 0.10 & 0.12 \\
Basal-unit horizons & $\{2, 3\}$ & 1 \\
Fault style & Single, raised, or lowered central block & Raised central block \\
Throw of fault pair & Shared, $\mathrm{Uniform}(0.02, 0.20)$ & Independent per fault, $\mathrm{Uniform}(0.02, 0.20)$ \\
Fault-pair width, $w$ & $\mathrm{Uniform}(0.15, 0.30)$ & $\mathrm{Uniform}(0.35, 0.50)$ \\
Tip elevation, $z_{\mathrm{top}}$ & $\mathrm{Uniform}(0.79, 1.19)$ & $\mathrm{Uniform}(0.80, 1.00)$ \\
Supplementary fault & Probability 0.30 & Always \\
\bottomrule
\end{tabularx}
\end{table}

The extended-range subset keeps the structural generator of Section \ref{sec:geometry_generation} but shifts several of its parameter ranges (Table \ref{tab:ood_1_stats}). The faulted sequence is built from six to eight horizons instead of three to five, with smaller amplitudes and higher frequencies, and the layers therefore become more numerous, thinner, and nearly flat. Each geometry contains 8 to 10 stratigraphic layers, compared with 6 to 9 in the main dataset. The fault configuration is narrowed to a single style. Every geometry contains a fault pair with a raised central block, together with a supplementary fault, a configuration found in only 7\% of the geometries in the main dataset. The fault pair is also wider than in the main dataset, with a width of 0.35 to 0.50 instead of 0.15 to 0.30, and its two faults have independent rather than shared throws.

\begin{figure}[htbp]
    \centering
    \includegraphics[width=\textwidth]{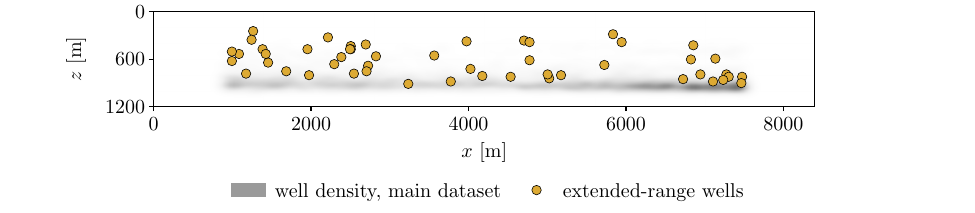}
    \caption{Injection-well locations of the extended-range subset, shown over the density of well locations in the main dataset. Coordinates as in Figure \ref{fig:well_placement_coverage}.}
    \label{fig:wells_extended}
\end{figure}

These changes carry through to the properties. Properties follow the classification and facies populations of Section \ref{sec:property_sampling}, but the additional layers raise the number of permeable layers to between 5 and 7 per property realisation, compared with 3 to 6 in the main dataset. Facies are assigned from the bottom of the sequence upward, so the extra layers receive higher-permeability populations, and the mean permeability of the permeable layers rises to 411.1\,mD, compared with 271.6\,mD. The proportion of sealing fault cores is also higher, at 45\% compared with 30.7\%. The wells, one to four per property realisation, lie at shallower depths than in the main dataset, with a median of 635\,m compared with 905\,m (Figure \ref{fig:wells_extended}).

\begin{figure}[htbp]
    \centering
    \includegraphics[width=\textwidth]{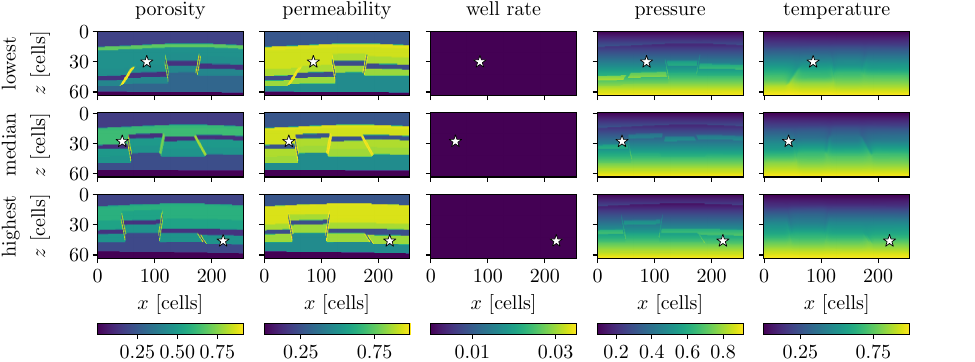}
    \caption{Input fields of the extended-range cases with the lowest, median, and highest plume-masked MAE averaged over reporting times (rows): porosity, permeability, the well mask, and the pressure and temperature fields at $t = 0$, with all fields except the well mask in normalised units. Axes as in Figure \ref{fig:case_overview}, and stars mark the injection wells.}
    \label{fig:inputs_extended}
\end{figure}

Figure \ref{fig:inputs_extended} shows the input fields of the extended-range cases with the lowest, median, and highest plume-masked MAE, discussed in Section \ref{sec:surrogate_extrapolation_results}. The thin, nearly flat layers and the wide fault pair are visible in the porosity and permeability fields of all three cases.

\subsection{Parallel-fault subset}

The parallel-fault subset departs further from the training data, since it is built with a separate generator that produces a different fault topology (Table \ref{tab:ood_2_stats}). Two to four faults with a common dip cut a faulted tier (the faulted region of Section~\ref{sec:surrogate_extrapolation_results}), each with its own throw and downthrown side, and the tier is therefore divided into a sequence of blocks with independent offsets. In the main dataset, by contrast, faults form either a single fault or a pair bounding one central block. Growth-wedge layers above and below the tier absorb the offsets, and unfaulted base and cap tiers complete the geometry, with the faulted tier placed low in the domain in 8 geometries and high in the remaining 10. Each geometry contains 9 to 13 stratigraphic layers, compared with 6 to 9 in the main dataset. The throws, between 0.02 and 0.08, lie within the lower part of the 0.02 to 0.20 range of the main fault pairs. The common dip of 55--65$^\circ$ in normalised coordinates corresponds to 25--36$^\circ$ in the physical domain, shallower than the main faults in the main dataset but similar to the supplementary faults. Generation is seeded by the geometry index and is therefore reproducible.

\begin{table}[htbp]
\centering
\caption{Generator parameters for the parallel-fault subset. Where two values are given, they correspond to the low and high positions of the faulted tier.}
\label{tab:ood_2_stats}
\small
\renewcommand{\arraystretch}{1.15}
\begin{tabularx}{\linewidth}{>{\raggedright\arraybackslash}p{0.36\linewidth} >{\raggedright\arraybackslash}X}
\toprule
Parameter & Distribution / value \\
\midrule
Number of faults & $\{2, 3, 4\}$ \\
Fault locations & $\mathrm{Uniform}(0.8, 2.0)$, at least 0.15 apart \\
Dip angle & $\mathrm{Uniform}(55^\circ, 65^\circ)$ in normalised coordinates (25--36$^\circ$ physical), shared by all faults; dip direction random \\
Throw & $\mathrm{Uniform}(0.02, 0.08)$, independently per fault \\
Downthrown side & Left or right, independently per fault \\
Damage-zone thickness & 0.05 \\
Fault-core width & $\mathrm{Uniform}(0.15, 0.60)$ of the damage-zone half-width \\
Layers in faulted tier & $\{3, 4, 5\}$ \\
Layers in base and cap tiers & $\{2, 3\}$ each, plus one growth-wedge layer below and above the faulted tier \\
Nominal extent of faulted tier & $(0.15, 0.55)$ or $(0.42, 0.90)$ \\
Top of base tier & 0.05 or 0.10 \\
Bottom of cap tier & 0.85 or 1.15 \\
Horizon positions in faulted tier & $\mathrm{Uniform}(0.1, 0.9)$ of tier height, at least 0.05 apart \\
Horizon amplitude & $\mathrm{Uniform}(0.05, 0.35)$ of half the smallest layer spacing \\
Horizon frequency & $\mathrm{Uniform}(0.2, 0.5)$ \\
Horizon phase & $\mathrm{Uniform}(-1, 1)$ \\
\bottomrule
\end{tabularx}
\end{table}

\begin{figure}[htbp]
    \centering
    \includegraphics[width=\textwidth]{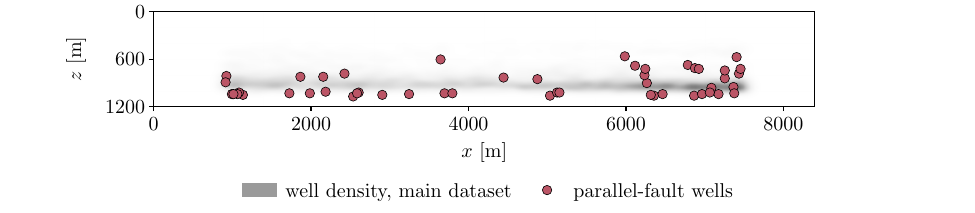}
    \caption{Injection-well locations of the parallel-fault subset, shown over the density of well locations in the main dataset, as in Figure \ref{fig:wells_extended}.}
    \label{fig:wells_parallel}
\end{figure}

As in the extended-range subset, the larger number of layers increases the number of permeable layers, here to between 6 and 10 per property realisation, compared with 3 to 6 in the main dataset, and raises their mean permeability to 552.6\,mD, compared with 271.6\,mD. Fault cores in this subset are assigned damage-zone properties, and no fault is sealing, whereas 30.7\% of fault cores are sealing in the main dataset. The wells, two or three per property realisation, cluster near the base of the permeable sequence, with a median depth of 1,025\,m compared with 905\,m (Figure \ref{fig:wells_parallel}).

\begin{figure}[htbp]
    \centering
    \includegraphics[width=\textwidth]{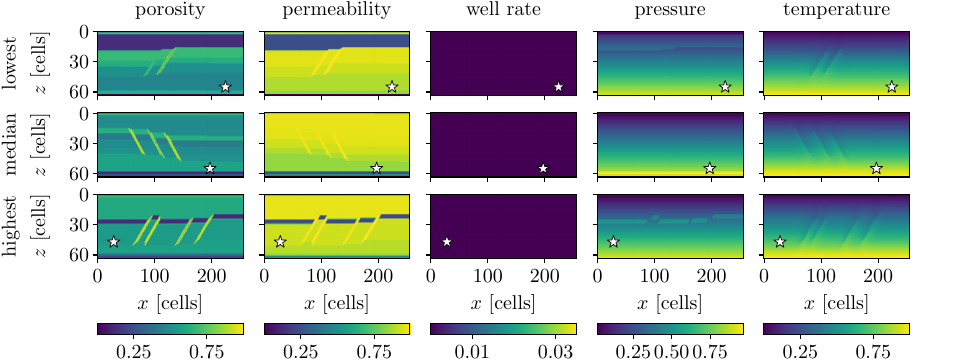}
    \caption{Input fields of the parallel-fault cases with the lowest, median, and highest plume-masked MAE averaged over reporting times (rows), as in Figure \ref{fig:inputs_extended}.}
    \label{fig:inputs_parallel}
\end{figure}

Figure \ref{fig:inputs_parallel} shows the input fields of the parallel-fault cases with the lowest, median, and highest plume-masked MAE. The parallel faults are visible in the porosity and permeability fields, and in the highest-error case the well lies beside the faulted region, beneath a thin low-permeability layer.

\clearpage

\bibliographystyle{plainnat}
\bibliography{references}

@article{pawar-2015,
  author  = {Pawar, Rajesh J. and Bromhal, Grant S. and Carey, J. William and Foxall, William and Korre, Anna and Ringrose, Philip S. and Tucker, Owain and Watson, Maxwell N. and White, Joshua A.},
  title   = {Recent Advances in Risk Assessment and Risk Management of Geologic {CO$_2$} Storage},
  journal = {International Journal of Greenhouse Gas Control},
  year    = {2015},
  volume  = {40},
  pages   = {292--311},
  doi     = {10.1016/j.ijggc.2015.06.014}
}

@article{bandilla-2015,
  author  = {Bandilla, Karl W. and Celia, Michael A. and Birkholzer, Jens T. and Cihan, Abdullah and Leister, Evan C.},
  title   = {Multiphase Modeling of Geologic Carbon Sequestration in Saline Aquifers},
  journal = {Groundwater},
  year    = {2015},
  volume  = {53},
  number  = {3},
  pages   = {362--377},
  doi     = {10.1111/gwat.12315}
}

@article{celia-2015,
  author  = {Celia, M. A. and Bachu, S. and Nordbotten, J. M. and Bandilla, K. W.},
  title   = {Status of {CO$_2$} Storage in Deep Saline Aquifers with Emphasis on Modeling Approaches and Practical Simulations},
  journal = {Water Resources Research},
  year    = {2015},
  volume  = {51},
  number  = {9},
  pages   = {6846--6892},
  doi     = {10.1002/2015WR017609}
}

@article{nordbotten-2024,
  author  = {Nordbotten, Jan M. and Fern{\o}, Martin A. and Flemisch, Bernd and Kovscek, Anthony R. and Lie, Knut-Andreas},
  title   = {The 11th {Society of Petroleum Engineers} Comparative Solution Project: Problem Definition},
  journal = {SPE Journal},
  year    = {2024},
  volume  = {29},
  number  = {5},
  pages   = {2507--2524},
  doi     = {10.2118/218015-PA}
}

@article{macdowell-2018,
  author  = {Bui, Mai and Adjiman, Claire S. and Bardow, Andr{\'e} and Anthony, Edward J. and Boston, Andy and Brown, Solomon and Fennell, Paul S. and Fuss, Sabine and Galindo, Amparo and Hackett, Leigh A. and Hallett, Jason P. and Herzog, Howard J. and Jackson, George and Kemper, Jasmin and Krevor, Samuel and Maitland, Geoffrey C. and Matuszewski, Michael and Metcalfe, Ian S. and Petit, Camille and Puxty, Graeme and Reimer, Jeffrey and Reiner, David M. and Rubin, Edward S. and Scott, Stuart A. and Shah, Nilay and Smit, Berend and Trusler, J. P. Martin and Webley, Paul and Wilcox, Jennifer and Mac Dowell, Niall},
  title   = {Carbon Capture and Storage ({CCS}): The Way Forward},
  journal = {Energy \& Environmental Science},
  year    = {2018},
  volume  = {11},
  number  = {5},
  pages   = {1062--1176},
  doi     = {10.1039/C7EE02342A}
}

@book{ipcc-2022,
  author    = {{IPCC}},
  title     = {Climate Change 2022: Mitigation of Climate Change. Contribution of Working Group {III} to the Sixth Assessment Report of the Intergovernmental Panel on Climate Change},
  publisher = {Cambridge University Press},
  address   = {Cambridge, UK and New York, NY, USA},
  year      = {2022},
  note      = {Edited by P.~R. Shukla, J.~Skea, R.~Slade, A.~Al~Khourdajie, R.~van~Diemen, D.~McCollum, M.~Pathak, S.~Some, P.~Vyas, R.~Fradera, M.~Belkacemi, A.~Hasija, G.~Lisboa, S.~Luz, and J.~Malley},
  doi       = {10.1017/9781009157926}
}

@book{metz-2005,
  editor    = {Metz, Bert and Davidson, Ogunlade and de Coninck, Heleen and Loos, Manuela and Meyer, Leo},
  title     = {{IPCC} Special Report on Carbon Dioxide Capture and Storage},
  publisher = {Cambridge University Press},
  address   = {Cambridge, UK},
  year      = {2005},
  url       = {https://www.ipcc.ch/report/carbon-dioxide-capture-and-storage/}
}

@article{ju-2024,
  author  = {Ju, Xin and Hamon, Fran{\c{c}}ois P. and Wen, Gege and Kanfar, Rayan and Araya-Polo, Mauricio and Tchelepi, Hamdi A.},
  title   = {Learning {CO$_2$} Plume Migration in Faulted Reservoirs with Graph Neural Networks},
  journal = {Computers \& Geosciences},
  year    = {2024},
  volume  = {193},
  pages   = {105711},
  doi     = {10.1016/j.cageo.2024.105711}
}

@article{wen-2022,
  author  = {Wen, Gege and Li, Zongyi and Azizzadenesheli, Kamyar and Anandkumar, Anima and Benson, Sally M.},
  title   = {{U-FNO}---An Enhanced {Fourier} Neural Operator-Based Deep-Learning Model for Multiphase Flow},
  journal = {Advances in Water Resources},
  year    = {2022},
  volume  = {163},
  pages   = {104180},
  doi     = {10.1016/j.advwatres.2022.104180}
}

@article{wen-2023,
  author  = {Wen, Gege and Li, Zongyi and Long, Qirui and Azizzadenesheli, Kamyar and Anandkumar, Anima and Benson, Sally M.},
  title   = {Real-Time High-Resolution {CO$_2$} Geological Storage Prediction Using Nested {Fourier} Neural Operators},
  journal = {Energy \& Environmental Science},
  year    = {2023},
  volume  = {16},
  number  = {4},
  pages   = {1732--1741},
  doi     = {10.1039/D2EE04204E}
}

@article{badawi-2025,
  author  = {Badawi, Daniel and Gildin, Eduardo},
  title   = {Neural Operator-Based Proxy for Reservoir Simulations Considering Varying Well Settings, Locations, and Permeability Fields},
  journal = {Computers \& Geosciences},
  year    = {2025},
  volume  = {196},
  pages   = {105826},
  doi     = {10.1016/j.cageo.2024.105826}
}

@article{lu-2024,
  author  = {Lu, Hannah and Sal{\'o}-Salgado, Llu{\'i}s and Marzouk, Youssef M. and Juanes, Ruben},
  title   = {Uncertainty Quantification of Fluid Leakage and Fault Instability in Geologic {CO$_2$} Storage},
  journal = {Water Resources Research},
  year    = {2025},
  volume  = {61},
  number  = {10},
  pages   = {e2024WR039275},
  doi     = {10.1029/2024WR039275}
}

@inproceedings{shirali-2023,
  author        = {Shirali, Ali and Abebe, Rediet and Hardt, Moritz},
  title         = {A Theory of Dynamic Benchmarks},
  booktitle     = {The Eleventh International Conference on Learning Representations ({ICLR})},
  year          = {2023},
  eprint        = {2210.03165},
  archivePrefix = {arXiv},
  primaryClass  = {cs.LG},
  url           = {https://openreview.net/forum?id=i8L9qoeZOS}
}

@article{egele-2024,
  author        = {Egele, Romain and Jacques Junior, J{\'u}lio C. S. and van Rijn, Jan N. and Guyon, Isabelle and Bar{\'o}, Xavier and Clap{\'e}s, Albert and Balaprakash, Prasanna and Escalera, Sergio and Moeslund, Thomas B. and Wan, Jun},
  title         = {{AI} Competitions and Benchmarks: Dataset Development},
  journal       = {arXiv preprint arXiv:2404.09703},
  year          = {2024},
  eprint        = {2404.09703},
  archivePrefix = {arXiv},
  primaryClass  = {cs.LG},
  doi           = {10.48550/arXiv.2404.09703},
  note          = {Preprint of Chapter 3 of \emph{Competitions and Benchmarks: The Science Behind the Contests}}
}

@article{bond-2015,
  author  = {Bond, Clare E.},
  title   = {Uncertainty in Structural Interpretation: Lessons to Be Learnt},
  journal = {Journal of Structural Geology},
  year    = {2015},
  volume  = {74},
  pages   = {185--200},
  doi     = {10.1016/j.jsg.2015.03.003}
}

@article{bond-2007,
  author  = {Bond, C. E. and Gibbs, A. D. and Shipton, Z. K. and Jones, S.},
  title   = {What Do You Think This Is? ``{C}onceptual Uncertainty'' in Geoscience Interpretation},
  journal = {GSA Today},
  year    = {2007},
  volume  = {17},
  number  = {11},
  pages   = {4--10},
  doi     = {10.1130/GSAT01711A.1}
}

@article{fluidflower-2024,
  author  = {Flemisch, Bernd and Nordbotten, Jan M. and Fern{\o}, Martin and Juanes, Ruben and Both, Jakub W. and Class, Holger and Delshad, Mojdeh and Doster, Florian and Ennis-King, Jonathan and Franc, Jacques and Geiger, Sebastian and Gl{\"a}ser, Dennis and Green, Christopher and Gunning, James and Hajibeygi, Hadi and Jackson, Samuel J. and Jammoul, Mohamad and Karra, Satish and Li, Jiawei and Matth{\"a}i, Stephan K. and Miller, Terry and Shao, Qi and Spurin, Catherine and Stauffer, Philip and Tchelepi, Hamdi and Tian, Xiaoming and Viswanathan, Hari and Voskov, Denis and Wang, Yuhang and Wapperom, Michiel and Wheeler, Mary F. and Wilkins, Andrew and Youssef, AbdAllah A. and Zhang, Ziliang},
  title   = {The {FluidFlower} Validation Benchmark Study for the Storage of {CO$_2$}},
  journal = {Transport in Porous Media},
  year    = {2024},
  volume  = {151},
  number  = {5},
  pages   = {865--912},
  doi     = {10.1007/s11242-023-01977-7}
}

@article{rasmussen-2020,
  author  = {Rasmussen, Atgeirr Fl{\o} and Sandve, Tor Harald and Bao, Kai and Lauser, Andreas and Hove, Joakim and Skaflestad, B{\aa}rd and Kl{\"o}fkorn, Robert and Blatt, Markus and Rustad, Alf Birger and S{\ae}vareid, Ove and Lie, Knut-Andreas and Thune, Andreas},
  title   = {The {Open Porous Media Flow} Reservoir Simulator},
  journal = {Computers \& Mathematics with Applications},
  year    = {2021},
  volume  = {81},
  pages   = {159--185},
  doi     = {10.1016/j.camwa.2020.05.014}
}

@incollection{wellmann-2018,
  author    = {Wellmann, Florian and Caumon, Guillaume},
  title     = {{3-D} Structural Geological Models: Concepts, Methods, and Uncertainties},
  booktitle = {Advances in Geophysics},
  publisher = {Elsevier},
  year      = {2018},
  volume    = {59},
  pages     = {1--121},
  doi       = {10.1016/bs.agph.2018.09.001}
}

@article{jayasekara-2020,
  author  = {Jayasekara, D. W. and Ranjith, P. G. and Wanniarachchi, W. A. M. and Rathnaweera, T. D.},
  title   = {Understanding the Chemico-Mineralogical Changes of Caprock Sealing in Deep Saline {CO$_2$} Sequestration Environments: A Review Study},
  journal = {The Journal of Supercritical Fluids},
  year    = {2020},
  volume  = {161},
  pages   = {104819},
  doi     = {10.1016/j.supflu.2020.104819}
}

@article{sun-2020,
  author  = {Sun, Qian and Ampomah, William and Kutsienyo, Eusebius Junior and Appold, Martin and Adu-Gyamfi, Benjamin and Dai, Zhenxue and Soltanian, Mohamed Reza},
  title   = {Assessment of {CO$_2$} Trapping Mechanisms in Partially Depleted Oil-Bearing Sands},
  journal = {Fuel},
  year    = {2020},
  volume  = {278},
  pages   = {118356},
  doi     = {10.1016/j.fuel.2020.118356}
}

@article{luo-2022,
  author  = {Luo, A. and Li, Y. M. and Chen, X. and Zhu, Z. Y. and Peng, Y.},
  title   = {Review of {CO$_2$} Sequestration Mechanism in Saline Aquifers},
  journal = {Natural Gas Industry B},
  year    = {2022},
  volume  = {9},
  number  = {4},
  pages   = {383--393},
  doi     = {10.1016/j.ngib.2022.07.002}
}

@article{wu-2021,
  author  = {Wu, Hao and Jayne, Richard S. and Bodnar, Robert J. and Pollyea, Ryan M.},
  title   = {Simulation of {CO$_2$} Mineral Trapping and Permeability Alteration in Fractured Basalt: Implications for Geologic Carbon Sequestration in Mafic Reservoirs},
  journal = {International Journal of Greenhouse Gas Control},
  year    = {2021},
  volume  = {109},
  pages   = {103383},
  doi     = {10.1016/j.ijggc.2021.103383}
}

@article{vrolijk-2016,
  author  = {Vrolijk, Peter J. and Urai, Janos L. and Kettermann, Michael},
  title   = {Clay Smear: Review of Mechanisms and Applications},
  journal = {Journal of Structural Geology},
  year    = {2016},
  volume  = {86},
  pages   = {95--152},
  doi     = {10.1016/j.jsg.2015.09.006}
}

@article{salo-salgado-2023,
  author  = {Sal{\'o}-Salgado, Llu{\'i}s and Davis, Steven and Juanes, Ruben},
  title   = {Fault Permeability from Stochastic Modeling of Clay Smears},
  journal = {Geology},
  year    = {2023},
  volume  = {51},
  number  = {1},
  pages   = {91--95},
  doi     = {10.1130/G50739.1}
}

@article{tacher-2006,
  author  = {Tacher, Laurent and Pomian-Srzednicki, Izabela and Parriaux, Aur{\`e}le},
  title   = {Geological Uncertainties Associated with {3-D} Subsurface Models},
  journal = {Computers \& Geosciences},
  year    = {2006},
  volume  = {32},
  number  = {2},
  pages   = {212--221},
  doi     = {10.1016/j.cageo.2005.06.010}
}

@article{wellmann-2010,
  author  = {Wellmann, J. Florian and Horowitz, Franklin G. and Schill, Eva and Regenauer-Lieb, Klaus},
  title   = {Towards Incorporating Uncertainty of Structural Data in {3D} Geological Inversion},
  journal = {Tectonophysics},
  year    = {2010},
  volume  = {490},
  number  = {3--4},
  pages   = {141--151},
  doi     = {10.1016/j.tecto.2010.04.022}
}

@inproceedings{charles-2001,
  author       = {Charles, T. and Gu{\'e}m{\'e}n{\'e}, J.-M. and Corre, B. and Vincent, G. and Dubrule, O.},
  title        = {Experience with the Quantification of Subsurface Uncertainties},
  booktitle    = {SPE Asia Pacific Oil and Gas Conference and Exhibition},
  year         = {2001},
  address      = {Jakarta, Indonesia},
  organization = {Society of Petroleum Engineers},
  note         = {Paper SPE-68703-MS},
  doi          = {10.2118/68703-MS}
}

@article{ghyselincks-2025,
  author  = {Ghyselincks, Simon and Okhmak, Valeriia and Zampini, Stefano and Turkiyyah, George and Keyes, David and Haber, Eldad},
  title   = {Synthetic Geology: Structural Geology Meets Deep Learning},
  journal = {Journal of Geophysical Research: Machine Learning and Computation},
  year    = {2026},
  volume  = {3},
  number  = {1},
  pages   = {e2025JH000986},
  doi     = {10.1029/2025JH000986}
}

@article{abdellatif-2026,
  author  = {Abdellatif, Alhasan and Menke, Hannah P. and Maes, Julien and Elsheikh, Ahmed H. and Doster, Florian},
  title   = {A Benchmark Dataset for Machine Learning Surrogates of Pore-Scale {CO$_2$}--Water Interaction},
  journal = {Scientific Data},
  year    = {2026},
  volume  = {13},
  pages   = {621},
  doi     = {10.1038/s41597-025-05794-z}
}

@article{landa-marban-2025,
  author  = {Landa-Marb{\'a}n, David and Sandve, Tor H.},
  title   = {pyopmspe11: A {Python} Framework Using {OPM Flow} for the {SPE11} Benchmark Project},
  journal = {Journal of Open Source Software},
  year    = {2025},
  volume  = {10},
  number  = {105},
  pages   = {7357},
  doi     = {10.21105/joss.07357}
}

@inproceedings{ronneberger-2015,
  author        = {Ronneberger, Olaf and Fischer, Philipp and Brox, Thomas},
  title         = {{U-Net}: Convolutional Networks for Biomedical Image Segmentation},
  booktitle     = {Medical Image Computing and Computer-Assisted Intervention -- {MICCAI} 2015},
  series        = {Lecture Notes in Computer Science},
  volume        = {9351},
  year          = {2015},
  pages         = {234--241},
  publisher     = {Springer},
  doi           = {10.1007/978-3-319-24574-4_28},
  eprint        = {1505.04597},
  archivePrefix = {arXiv}
}

@inproceedings{kingma-2014,
  author        = {Kingma, Diederik P. and Ba, Jimmy},
  title         = {Adam: A Method for Stochastic Optimization},
  booktitle     = {3rd International Conference on Learning Representations ({ICLR})},
  year          = {2015},
  eprint        = {1412.6980},
  archivePrefix = {arXiv},
  primaryClass  = {cs.LG},
  url           = {https://arxiv.org/abs/1412.6980}
}

@inproceedings{loshchilov-2019,
  author        = {Loshchilov, Ilya and Hutter, Frank},
  title         = {Decoupled Weight Decay Regularization},
  booktitle     = {7th International Conference on Learning Representations ({ICLR})},
  year          = {2019},
  eprint        = {1711.05101},
  archivePrefix = {arXiv},
  primaryClass  = {cs.LG},
  url           = {https://openreview.net/forum?id=Bkg6RiCqY7}
}

@inproceedings{wu-2018,
  author        = {Wu, Yuxin and He, Kaiming},
  title         = {Group Normalization},
  booktitle     = {Computer Vision -- {ECCV} 2018},
  series        = {Lecture Notes in Computer Science},
  volume        = {11217},
  pages         = {3--19},
  publisher     = {Springer},
  year          = {2018},
  eprint        = {1803.08494},
  archivePrefix = {arXiv},
  url           = {https://arxiv.org/abs/1803.08494}
}

@article{geuzaine-2009,
  author  = {Geuzaine, Christophe and Remacle, Jean-Fran{\c{c}}ois},
  title   = {{Gmsh}: A 3-{D} Finite Element Mesh Generator with Built-in Pre- and Post-Processing Facilities},
  journal = {International Journal for Numerical Methods in Engineering},
  year    = {2009},
  volume  = {79},
  number  = {11},
  pages   = {1309--1331},
  doi     = {10.1002/nme.2579}
}

@book{fossen-2016,
  author    = {Fossen, Haakon},
  title     = {Structural Geology},
  edition   = {2nd},
  publisher = {Cambridge University Press},
  address   = {Cambridge, UK},
  year      = {2016}
}

@article{abdoulghafour-2019,
  author  = {Abdoulghafour, Halidi and Sarmadivaleh, Mohammad and Hauge, Lars Petter and Fern{\o}, Martin and Iglauer, Stefan},
  title   = {Capillary Pressure Characteristics of {CO$_2$}--Brine--Sandstone Systems},
  journal = {International Journal of Greenhouse Gas Control},
  year    = {2020},
  volume  = {94},
  pages   = {102876},
  doi     = {10.1016/j.ijggc.2019.102876}
}

@article{jessell-2022,
  author  = {Jessell, M. and Guo, J. and Li, Y. and Lindsay, M. and Scalzo, R. and Giraud, J. and Pirot, G. and Cripps, E. and Ogarko, V.},
  title   = {Into the {Noddyverse}: A Massive Data Store of {3D} Geological Models for Machine Learning and Inversion Applications},
  journal = {Earth System Science Data},
  year    = {2022},
  volume  = {14},
  number  = {1},
  pages   = {381--392},
  doi     = {10.5194/essd-14-381-2022}
}
\end{document}